\documentclass[10pt,twocolumn,letterpaper]{article}
\usepackage{cvpr}
\usepackage[dvipsnames]{xcolor}
\usepackage{multirow}
\usepackage{acronym}
\usepackage{colortbl}
\usepackage{graphicx} \graphicspath{{figures/}}
\usepackage{bm}
\usepackage{lipsum}
\usepackage{float}
\usepackage{algorithmic}
\usepackage[linesnumbered,ruled,vlined]{algorithm2e}
\usepackage[misc]{ifsym}
\usepackage{dblfloatfix}
\usepackage{mdframed}
\usepackage{listings}
\usepackage{tcolorbox}
\usepackage{overpic}

\tcbuselibrary{breakable}
\newtcolorbox{breakbox}[1][]{
  breakable,
  colback=green!2,
  colframe=black!50,
  title=Prompt for pre-trained controllers,
  width=\columnwidth,
  #1
}
\newtcolorbox{breakbox1}[1][]{
  breakable,
  colback=yellow!2,
  colframe=black!50,
  title=Reward generated by GPT-o1-preview,
  width=\columnwidth,
  #1
}
\newtcolorbox{breakbox2}[1][]{
  breakable,
  colback=yellow!4,
  colframe=black!50,
  title=Reward generated by GPT-4o,
  width=\columnwidth,
  #1
}

\newtcolorbox{breakbox3}[1][]{
  breakable,
  colback=yellow!4,
  colframe=black!50,
  title=Reward generated by Gemini-1.5-Pro,
  width=\columnwidth,
  #1
}

\usepackage{xspace}
\usepackage{setspace}
\usepackage[skip=3pt,font=small]{caption}
\makeatletter
\renewcommand{\paragraph}{%
  \@startsection{paragraph}{4}%
  {\z@}{1ex \@plus 1ex \@minus .2ex}{-1em}%
  {\normalfont\normalsize\bfseries}%
}
\makeatother

\definecolor{cvprblue}{rgb}{0.21,0.49,0.74}
\definecolor{LightCyan}{rgb}{0.95,1,1}
\definecolor{myred}{rgb}{0.86, 0.17, 0.23}
\definecolor{orange}{RGB}{204,132,59}
\definecolor{blue}{RGB}{80,110,128}

\acrodef{ai}[AI]{Artificial Intelligence}
\acrodef{rl}[RL]{Reinforcement Learning}
\acrodef{llm}[LLM]{Large Language Model}
\acrodef{vlm}[VLM]{Vision Language Model}
\acrodef{rdp}[RDP]{Reward Design Problem}

\def\model{\texttt{GROVE}\xspace}
\def\posetoclip{\texttt{Pose2CLIP}\xspace}
\def\clipmodel{\texttt{CLIP-ViT-B/32}\xspace}
\definecolor{cvprblue}{rgb}{0.21,0.49,0.74}
\usepackage[pagebackref,breaklinks,colorlinks,allcolors=cvprblue,bookmarks=false]{hyperref}

\title{\model{}: A Generalized Reward for Learning Open-Vocabulary Physical Skill}
\author{
    Jieming Cui$^{1,2*}$, Tengyu Liu$^{2*}$, Ziyu Meng$^{2,3}$, Jiale Yu$^{4}$, Ran Song$^{3}$, Wei Zhang$^{3}$,\vspace{3pt}\\
    Yixin Zhu$^{1,~\textrm{\Letter}}$, Siyuan Huang$^{2,~\textrm{\Letter}}$
    \vspace{6pt}\\
    \small $^1$ Institute for Artificial Intelligence, Peking University\quad{}
    $^2$ State Key Laboratory of General Artificial Intelligence, BIGAI\\
    \small $^3$ School of Control Science and Engineering, Shandong University
    \small $^4$ Department of Automation, Tsinghua University
    \vspace{6pt}\\
    \href{https://jiemingcui.github.io/grove/}{https://jiemingcui.github.io/grove/}
}

\begin{document}

\twocolumn[{
\renewcommand\twocolumn[1][]{#1}
\maketitle
\begin{center}
    \small
    \captionsetup{type=figure}
    \includegraphics[width=\linewidth]{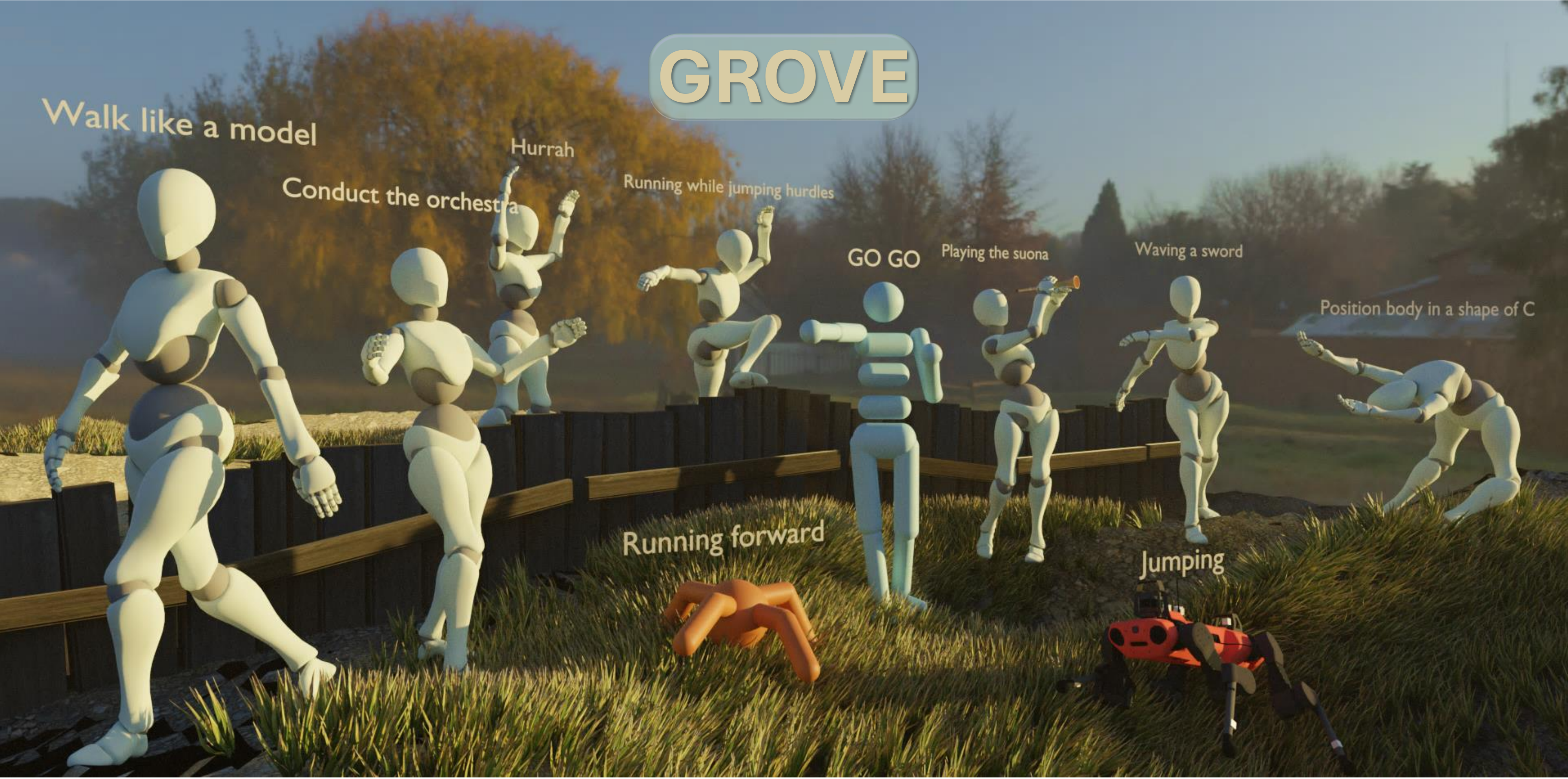}
    \captionof{figure}{\textbf{Open-vocabulary physical skills learned with our generalized reward framework \model.} The white humanoid mannequins demonstrate diverse skills from abstract instructions (\eg, ``conduct the orchestra,'' ``position body in a shape of C'') using a pre-trained controller. The system also generalizes to standard \acs{rl} benchmarks, including the \textcolor{orange}{Ant}, \textcolor{Turquoise}{Humanoid}, and quadrupedal \textcolor{RedOrange}{ANYmal}, all without task-specific reward engineering.}
    \label{fig:teaser}
\end{center}
}]

\begin{abstract}
Learning open-vocabulary physical skills for simulated agents presents a significant challenge in \ac{ai}. Current \ac{rl} approaches face critical limitations: manually designed rewards lack scalability across diverse tasks, while demonstration-based methods struggle to generalize beyond their training distribution. We introduce \model, a generalized reward framework that enables open-vocabulary physical skill learning without manual engineering or task-specific demonstrations. Our key insight is that \acp{llm} and \acp{vlm} provide complementary guidance---\acp{llm} generate precise physical constraints capturing task requirements, while \acp{vlm} evaluate motion semantics and naturalness. Through an iterative design process, \ac{vlm}-based feedback continuously refines \ac{llm}-generated constraints, creating a self-improving reward system. To bridge the domain gap between simulation and natural images, we develop \posetoclip, a lightweight mapper that efficiently projects agent poses directly into semantic feature space without computationally expensive rendering. Extensive experiments across diverse embodiments and learning paradigms demonstrate \model's effectiveness, achieving {22.2\%} higher motion naturalness and {25.7\%} better task completion scores while training 8.4$\times$ faster than previous methods. These results establish a new foundation for scalable physical skill acquisition in simulated environments.
\end{abstract}

\section{Introduction}

Enabling simulated agents to learn diverse physical skills from natural language instructions represents a fundamental challenge in \ac{ai} for graphics and robotics. Despite significant advances in \ac{rl}, developing agents that perform arbitrary physical tasks specified through natural language remains difficult due to the traditional reliance on manually engineered reward functions~\cite{sutton2018reinforcement}. For example, training a humanoid to \textit{run forward} involves a complex reward system accounting for velocity, energy consumption, survival, and more~\cite{makoviychuk2021isaac}. This tailored approach limits the generalizability of \ac{rl} methods to open-vocabulary tasks.

The central challenge lies in developing generalizable reward mechanisms that can both interpret arbitrary natural language instructions and accurately evaluate complex motion patterns. While imitation learning methods offer partial generalizability~\cite{juravsky2022padl,juravsky2024superpadl,tessler2024maskedmimic}, they remain fundamentally constrained by their text-motion training distributions. Recent approaches have made progress by leveraging \acp{llm}~\cite{achiam2023gpt,touvron2023llama,anil2023palm} and \acp{vlm}~\cite{radford2021learning,alayrac2022flamingo} for automatic reward design. However, they each face distinct limitations:

\textbf{\ac{llm}-based approches} excel at generating precise, dynamic constraints from natural language instructions~\cite{ma2024eureka,han2024autoreward}, providing detailed reward signals for specific kinematic aspects such as joint positions, velocities, and spatial relationships. However, they struggle with holistic motion assessment, focusing on individual constraints while missing stylistic qualities and natural movement patterns. This limitation frequently results in technically correct but unnatural motions requiring substantial human refinement~\cite{son2024llm,wang2024prompt,sun2024large}. Conversely, \textbf{\ac{vlm}-based approaches} provide rich semantic feedback through visual evaluation~\cite{rocamonde2024vlmrm,cui2024anyskill}, effectively assessing whether actions ``look correct'' perceptually. While offering the robust evaluation of stylistic correctness and naturalness that \acp{llm} typically miss, these approaches struggle with temporal consistency across motion sequences. They cannot enforce the precise kinematic constraints necessary for skill mastery.

We propose a generalized reward framework, \model, that leverages the complementary strengths of both sides. By combining \ac{llm}-generated constraints with \ac{vlm}-based semantic evaluation,  \model enables robust open-vocabulary physical skill learning. Intuitively, these two reward sources provide orthogonal constraints on the \ac{rl} search space: \ac{llm}-generated rewards impose precise physical and kinematic constraints, while \ac{vlm}-based rewards constrain the solutions to semantically valid and natural-looking motions. This orthogonality ensures the combined reward effectively narrows the search space to solutions satisfying physical accuracy and semantic correctness. 

A significant technical challenge is bridging the visual domain gap between simplified simulation environments and the natural images on which \acp{vlm} are trained. We address this with \posetoclip, a lightweight model that efficiently maps agent poses directly to CLIP feature space without rendering. Trained on 1.7 million frames of high-quality rendered human poses, this mapping enables efficient semantic evaluation while preserving CLIP's rich representational capabilities and dramatically reducing computational requirements during training.

Extensive experiments demonstrate \model's effectiveness across diverse embodiments and tasks in two distinct settings: (i) integrating with hierarchical controllers pre-trained on human motion datasets and (ii) guiding standard \ac{rl} algorithms learning from scratch; see also \cref{fig:teaser}. Our evaluations show that combining \ac{llm}-generated constraints with \ac{vlm}-based feedback converges 8.4x faster and produces more natural movements than using either reward source independently. User studies confirm these improvements, showing a 22.2\% enhancement in perceived motion naturalness and a 25.7\% increase in task completion scores compared to baseline methods.

To summarize, our contributions are four-fold:
\begin{itemize}[leftmargin=*,nolistsep,noitemsep]
    \item We introduce \model, a generalized reward framework that combines \ac{llm}-generated precise constraints with \ac{vlm}-based semantic evaluation, enabling robust open-vocabulary physical skill learning without manual reward design or task-specific demonstrations.
    \item We develop \posetoclip, a lightweight pose-to-semantic feature mapper that bridges the domain gap between simulation and natural images, enabling efficient and effective semantic evaluation.
    \item We show \model's capability to generate diverse and natural motions for arbitrary embodiments from open-vocabulary instructions. Through extensive experiments on complex humanoid tasks and standard \ac{rl} benchmarks, our method achieves 25.7\% higher task completion score, 22.2\% improved motion naturalness, and $8.4\times$ faster convergence compared to baselines.
    \item We demonstrate the generalizability of our proposed reward across multiple scenarios, from leveraging pre-trained controllers to learning from scratch, and provide comprehensive ablation studies showing the complementary benefits of combining \ac{llm} and \ac{vlm} rewards.
\end{itemize}

\section{Related Work}

\paragraph{Physical skill learning and reward design}

Research in physical skill learning has followed two primary approaches to address the challenges of reward design. Traditional \ac{rl} methods rely on carefully engineered reward functions decomposed into multiple weighted components~\cite{peng2018deepmimic,andrychowicz2020learning,rajeswaran2017learning,li2024reinforcement,yu2018learning,li2021reinforcement,song2021deep,abdolhosseini2019learning}. While these approaches achieve impressive results for specific tasks, their dependence on manual tuning and domain expertise limits scalability to new behaviors.

Adversarial imitation learning offers an alternative by learning reward functions directly from demonstrations through discriminator networks~\cite{tessler2023calm,juravsky2024superpadl,liu2024learning,peng2021amp,peng2022ase,yao2022controlvae,luo2023perpetual,juravsky2022padl}. These approaches capture complex motion qualities without explicit reward engineering but are constrained by their reliance on extensive demonstration data and limited generalization beyond the training distribution. In contrast, our work leverages foundation models to provide complementary supervision signals that eliminate the need for both manual reward design and task-specific demonstrations.

\paragraph{Language-guided skill learning}

Foundation models have significant advanced zero-shot skill learning through two complementary approaches: \ac{llm}-based reward generation and \ac{vlm}-based semantic evaluation.

\ac{llm}-based methods~\cite{yu2023language,xie2024text2reward,ma2024eureka,ma2024dreureka} translate natural language instructions into programmatic reward functions by prompting language models to formulate physical constraints and success criteria. 
Recent work like Eureka~\cite{ma2024eureka} demonstrated that models such as GPT can generate precise rewards for controlling joint positions, velocities, and spatial relationships without human intervention. However, these approaches typically lack a holistic understanding of motion quality and cannot visually assess complex movement patterns. 
\ac{vlm}-based methods~\cite{rocamonde2024vlmrm,cui2024anyskill} use vision-language models to provide semantic feedback on generated motions. These approaches excel at determining whether actions appear natural and contextually appropriate. However, they face technical barriers including the substantial domain gap between simplified simulation renderings and natural images, challenges in maintaining temporal consistency across frames, and the inability to enforce precise physical constraints. Our approach, \posetoclip, bridges the domain gap while combining the strengths of both model types---\acp{llm} for precise constraint formulation and \acp{vlm} for holistic semantic evaluation.

\paragraph{Alternative approaches to zero-shot skill acquisition}

Beyond reward-based approaches, recent work explores alternative paradigms for zero-shot physical skill acquisition. Representation learning methods~\cite{tessler2024maskedmimic,luo2023universal} focus on building task-agnostic motion embeddings from demonstration data that capture underlying motion structure. For example, MaskedMimic~\cite{tessler2024maskedmimic} achieves zero-shot generalization by training models to reconstruct strategically masked portions of motion sequences. However, these approaches struggle to interpret open-vocabulary language instructions.

Code generators~\cite{liang2023code,huang2023voxposer} use \acp{llm} to translate instructions directly into executable robot commands, excelling in structured environments with well-defined actions. However, they face substantial challenges with dynamic full-body control and complex physical interactions. They primarily succeed in scenarios with inherent physical stability (\eg, table-mounted manipulators) and rely heavily on decomposing tasks into discrete, atomic actions. Our approach differs by creating a unified reward mechanism that handles both the precise kinematics and semantic correctness required for complex whole-body movements without requiring predefined action primitives.

\begin{figure}
    \small
    \centering	
    \includegraphics[width=\linewidth]{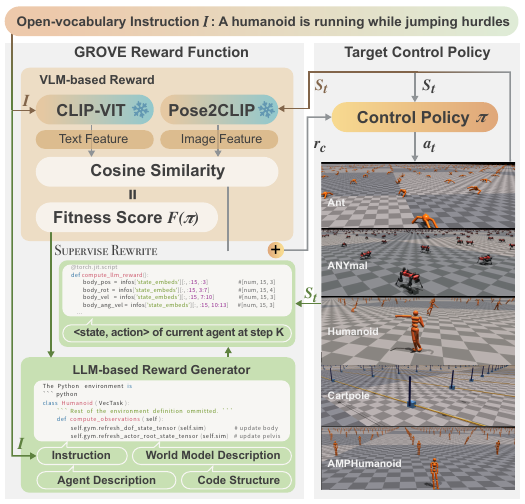}
    \caption{\textbf{The overall architecture of \model.} \model combines multiple components for zero-shot physical skill learning: a \acs{vlm}-based reward for semantic evaluation, a \posetoclip mapper to bridge the simulation-to-reality gap, an \acs{llm}-based reward generator for precise constraints, and an iterative reward design process. We demonstrate \model's effectiveness across five diverse agents: Ant, ANYmal, Humanoid, Cartpole, and AMPHumanoid.
     We demonstrate its effectiveness across five diverse agents: Ant, ANYmal, Humanoid, Cartpole, and AMPHumanoid.}
    \label{fig:model}
\end{figure}

\section{The \model{} Framework}

We introduce \model{}, a generalized reward framework that enables learning open-vocabulary physical skills without manual reward engineering or task-specific demonstrations. At its core, \model combines two complementary reward mechanisms: a \ac{vlm}-based component that evaluates the semantic correctness of performed actions, and an \ac{llm}-based component that formulates precise physical constraints. The framework uniquely leverages a \ac{vlm} to dynamically evaluate and re-sample rewards throughout the learning process (see \cref{fig:model}), creating a synergistic effect that addresses the limitations of each approach. The complete algorithmic implementation is detailed in \cref{alg:all}.

\begin{algorithm}[ht!]
    \footnotesize
    \SetAlgoLined 
    \caption{\model for physical skill learning{}}
         \label{alg:all}
    \KwIn{Task instruction $I$, world model $\xi$, agent description $D_\mathrm{agent}$, pre-trained \acs{vlm}, \acs{llm}}
    \textbf{Hyper-parameters:} {horzion length $K$} \\
    Initialize policy $\pi$; $R_L\leftarrow\acs{llm}(\xi, D_\mathrm{agent}, I)$\\
    \While{not converged}{
        Initialize $\mathcal{B}\leftarrow\emptyset$; $F(\pi) = \left[\hspace{1pt}\right]$; state $s$ \\
        \For{step $= 1,...,K$}{
            \texttt{\# Standard \acs{rl} operations} \\
            $a \leftarrow \pi(s)$, \, $s' \leftarrow \textsc{env}(s,a)$, \, $r\leftarrow R_\model(s',a,I)$ \\
            \texttt{\# Update fitness function} \\
            $F(\pi).\mathrm{append}(R_{V}(s',I))$\\
            \If{$\mathrm{8\ consecutive\ drops\ on\ F(\pi)}$ \& $\mathrm{F(\pi)_{last}\ < 0.1}$} {
                \texttt{\# Re-generate reward function} \\
                $R_L\leftarrow\acs{llm}(\xi, D_\mathrm{agent}, I)$\\
            }
            \texttt{\# Update policy and state variable} \\
            update $\mathcal{B}$ and $\pi$ according to PPO\\
            $s \leftarrow s'$
        }
    }
\end{algorithm}

\subsection{Generalized Reward}

We formalize \model as a \ac{rdp}~\cite{singh2009rewards}, building upon the framework introduced in Eureka~\cite{ma2024eureka}. An \ac{rdp} is defined as a tuple $P = \langle \xi, \mathcal{R}, F \rangle$, where $\mathcal{R}$ is the space of all possible reward functions, and $F$ is a fitness function that maps a policy $\pi$ to a real-valued fitness score $F(\pi)$. The world model $\xi = \langle S, A, T \rangle$ encapsulates the state space $S$, action space $A$, and transition function $T$ that governs the environment dynamics.

Given an arbitrary natural language instruction $I$, our key insight is that effective reward functions should combine both semantic understanding and precise physical constraints. We therefore design the \model reward as a weighted combination of complementary components:
\begin{equation}
    R_{\mathrm{GROVE}}(s,a,I) = \omega_{V}R_{V}(s,I) + \omega_{L}R_{L}(s,a;I),\label{eq:full_reward}
\end{equation}
where $R_{V}$ represents the \ac{vlm}-based semantic evaluation component, $R_{L}$ denotes the \ac{llm}-generated constraint component, and weights $\omega_{V}$ and $\omega_{L}$ balance their relative importance. Critically, $R_{\mathrm{GROVE}}$ is dynamically refined during training according to the fitness function $F$, enabling continuous improvement in instruction interpretation.

\paragraph{\ac{vlm}-based reward}
Our \ac{vlm}-based reward leverages a frozen CLIP model~\cite{radford2021learning} for semantic alignment between agent behaviors and natural language instructions. Traditionally, this would require computing CLIP similarity between rendered frames and instruction text---a computationally expensive process that often suffers from the sim-to-real gap. To address these limitations, we introduce \posetoclip, a specialized neural network that directly maps agent poses to the CLIP feature space without requiring explicit rendering. This design dramatically improves computational efficiency while also bridging the sim-to-real gap by training on high-fidelity Blender-rendered images that better match CLIP's training distribution of natural images. The \ac{vlm}-based reward is formally computed as:
\begin{equation}
    R_{V}(s,I) = \frac{
        \mathrm{CLIP}(I)\cdot\posetoclip(s)
    }{
        \Vert\mathrm{CLIP}(I)\Vert_2\cdot\Vert\posetoclip(s)\Vert_2
    }.
\end{equation}
A detailed analysis of \posetoclip's architecture and training methodology is provided in \cref{sec:method:pc}.

\paragraph{\ac{llm}-based reward} 

For the generation of task-specific reward functions $R_L$, we employ \textit{GPT-o1-preview}~\cite{achiam2023gpt} following the approach established in Eureka~\cite{ma2024eureka}. However, we introduce two critical enhancements to the prompt engineering process that improve reward quality. First, we enrich the context by incorporating detailed agent specifications ($D_{\mathrm{agent}}$), including precise joint nomenclature and indices that allow for more targeted reward formulation. Second, we explicitly guide the model with the instruction that auxiliary mechanisms already handle fundamental aspects like stability and locomotion, directing it to ``\textit{focus solely on capturing the essence of the task}.'' This strategic constraint prevents the model from generating overly complex rewards addressing already-solved problems. The generation and application of the \ac{llm}-based reward can be formalized as:
\begin{subequations}
    \begin{align}
        R_L(\cdot,\cdot\hspace{2pt};I) &= \acs{llm}(\xi, D_{\mathrm{agent}}, I), \label{eq:R_L} \\
        r_L &= R_L(s,a;I), \label{eq:r_l}
    \end{align}
\end{subequations}
where \cref{eq:R_L} is the generation of the reward function $R_L$, and \cref{eq:r_l} its evaluation on specific state-action pairs.

\paragraph{\ac{rdp} and fitness function} 

While \ac{llm}-based rewards offer detailed temporal awareness and precise constraint specification, they can suffer from instability and occasional semantic drift~\cite{ma2024eureka,li2024ag2manip}. To mitigate this limitation, we implement a quality control mechanism using our \ac{vlm}-based reward $R_V$ as a fitness evaluator for $R_L$. Specifically, we employ an adaptive rejection sampling strategy: the \ac{llm}-based reward function is regenerated whenever we detect a consistent performance decline, defined as 8 consecutive steps where the average $R_V$ across all parallel environments decreases relative to the previous step and the final average falls below a threshold (0.1). This verification ensures that optimization toward $R_L$ consistently results in visually and semantically correct behaviors, preventing the learning process from diverging toward unintended solutions that might satisfy the letter but not the spirit of the instruction.

\subsection{\posetoclip}\label{sec:method:pc}

A critical challenge in using \acp{vlm} for embodied control lies in the substantial domain gap between simulator-rendered images and the natural images on which models like CLIP were trained. This discrepancy undermines the effectiveness of direct CLIP-based reward computation. While high-fidelity rendering could theoretically bridge this gap, it introduces prohibitive computational overhead that makes real-time \ac{rl} impractical.

To address this challenge, we introduce \posetoclip, a lightweight neural network that directly maps agent pose representations to CLIP's feature space without requiring explicit rendering. This design serves dual purposes: it eliminates the computational burden of high-fidelity rendering (achieving a three-to-six-fold acceleration in our experiments) while preserving CLIP's semantic understanding capabilities. Though our implementation focuses on humanoid skill learning---arguably the most challenging embodiment scenario---the method generalizes to any agent with a fixed state space representation.

For training \posetoclip, we construct a diverse dataset combining SMPL~\cite{loper2015smpl} poses from the AMASS training split~\cite{mahmood2019amass} and the Motion-X dataset~\cite{lin2023motionx}. Each pose $\theta\in\mathbb{R}^{J\times3}$ represents the joint-wise Euler angles of the humanoid. To ensure comprehensive coverage of the pose distribution encountered during policy learning, we continuously augment this dataset with policy rollouts throughout the training process, ultimately yielding a refined dataset of 1.7 million frames after downsampling.

\begin{figure}[t!]
    \small
    \centering
    \begin{subfigure}[b]{0.495\linewidth}
        \includegraphics[width=\linewidth]{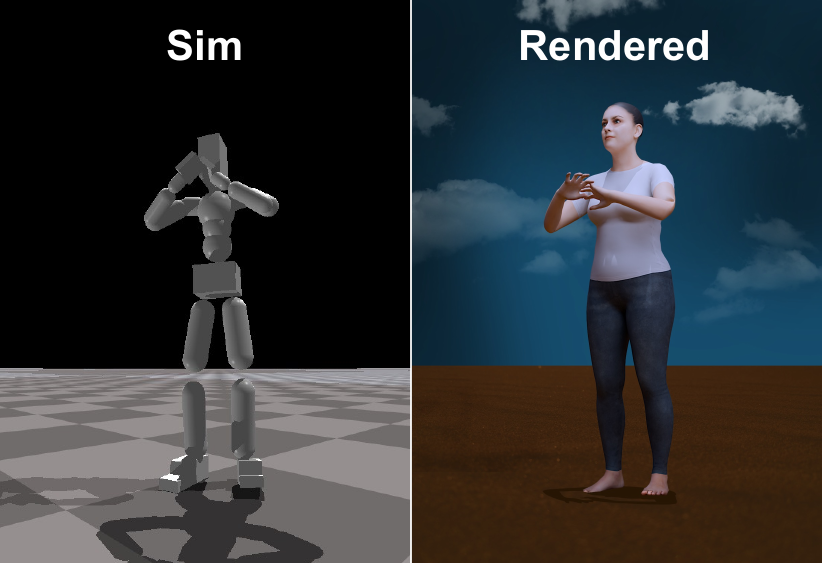}
        \caption{clap}
        \label{fig:render_clap}
    \end{subfigure}%
    \hfill
    \begin{subfigure}[b]{0.495\linewidth}
        \includegraphics[width=\linewidth]{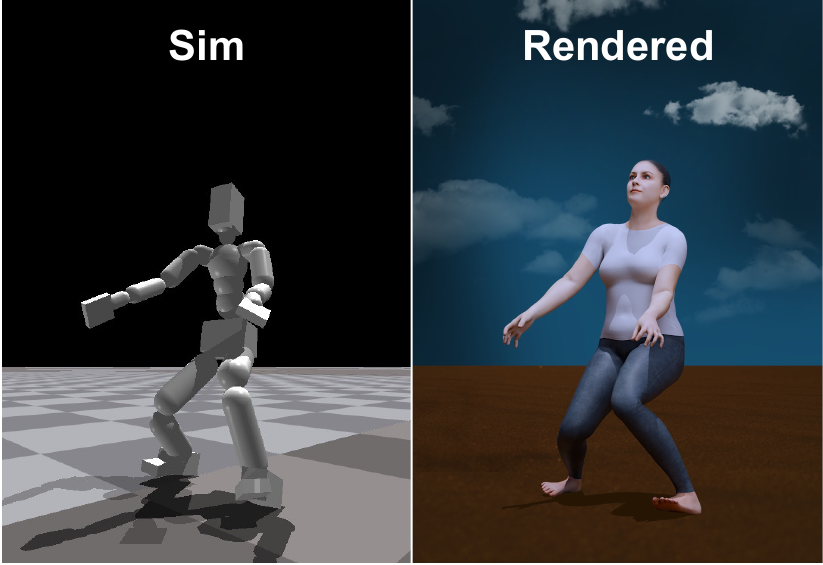}
        \caption{sit down}
        \label{fig:render_sit}
    \end{subfigure}%
    \caption{\textbf{Comparison of simulator vs. Blender-rendered images for \posetoclip training.} The left shows raw simulation, while the right the Blender renderings. \posetoclip maps agent poses directly to CLIP feature space, thus bridging the sim-to-real gap and improving reward quality for (a) ``clap'' and (b) ``sit down.''}
    \label{fig:render}
\end{figure}

The generation of ground truth CLIP features follows a carefully designed process. Using Blender with the SMPL-X Add-on~\cite{pavlakos2019expressive}, we render each pose with high-quality textures and realistic lighting. To minimize occlusion effects and ensure viewpoint robustness, we render each pose from five strategic angles (front, side, oblique, rear side, and rear) at 45° intervals (see \cref{fig:render}). These multi-view images are then processed through a pre-trained \clipmodel model~\cite{cherti2023reproducible} to obtain image feature embeddings, with features from different viewpoints of the same pose being mean-pooled to create a single representation.

The \posetoclip is intentionally lightweight---a two-layer MLP optimized to minimize MSE reconstruction loss between predicted and ground truth CLIP features. To counter the inherent data imbalance problem, where common poses dominate the dataset while important but rare poses are underrepresented, we implement a balanced sampling strategy. First, we apply k-means++~\cite{Arthur2007kmeansTA} to cluster the dataset into 500 semantically meaningful groups. Next, we employ a two-stage uniform sampling approach during training: we first sample uniformly across clusters to ensure the representation of all pose types, then sample uniformly within the selected clusters. This hierarchical sampling strategy ensures comprehensive coverage of the entire pose space, enhancing model generalization.

\begin{table}[t!]
    \centering
    \small
    \caption{\textbf{Quantitative evaluation of open-vocabulary humanoid skill synthesis.} We compare \model against state-of-the-art approaches in five metrics: task completion (C), motion Naturalness (N), motion smoothness (S), physical realism (P), and CLIP text-image similarity (CLIP\_S). Our approach integrating CALM outperforms alternatives in task completion while maintaining competitive or superior performance across other metrics.}
    \label{tab:comparison}
    \setlength{\tabcolsep}{3pt}
    \begin{tabular}{lcccccc}
        \toprule
        \textbf{}     & \textbf{C} $\uparrow$      & \textbf{N} $\uparrow$ & \textbf{S}$\downarrow$ & \textbf{P}$\uparrow$ & \textbf{CLIP\_S}$\uparrow$ \\
        \midrule
        AvatarCLIP~\cite{hong2022avatarclip}  &  1.621  &   4.793    &   0.411  & 5.821 & 22.105\\
        TMR~\cite{petrovich2023tmr}      &  3.793  &   5.966    &   \textbf{0.374}   & 7.684 & 18.885\\
        MoMask~\cite{guo2024momask}      &  3.621  &   6.724    &   0.492   & 7.863 & 22.372 \\
        MotionGPT~\cite{jiang2024motiongpt} &   5.207   &   6.552    &   0.885   & 7.000 & 23.142\\ 
        AnySkill~\cite{cui2024anyskill}      &   6.108    &   5.938    &  0.486    & 8.168 & 23.925\\ \midrule
        \textbf{Ours (+ CALM)} ~\cite{tessler2023calm}  &   \textbf{7.924}    &   \textbf{6.793}  &  0.488   & \textbf{8.452} & \textbf{28.998} \\
        \bottomrule
    \end{tabular}%
\end{table}

\section{Experiments}

Our experimental evaluation is structured in three parts: \cref{exp:pre} assesses \model's capability to generate open-vocabulary humanoid skills, \cref{exp:rl} evaluates its performance on standard \ac{rl} benchmarks across different embodiments, and \cref{sec:ablation} provides an ablation study demonstrating the contribution of each component.

\begin{figure*}[t!]
    \centering
    \includegraphics[width=\linewidth]{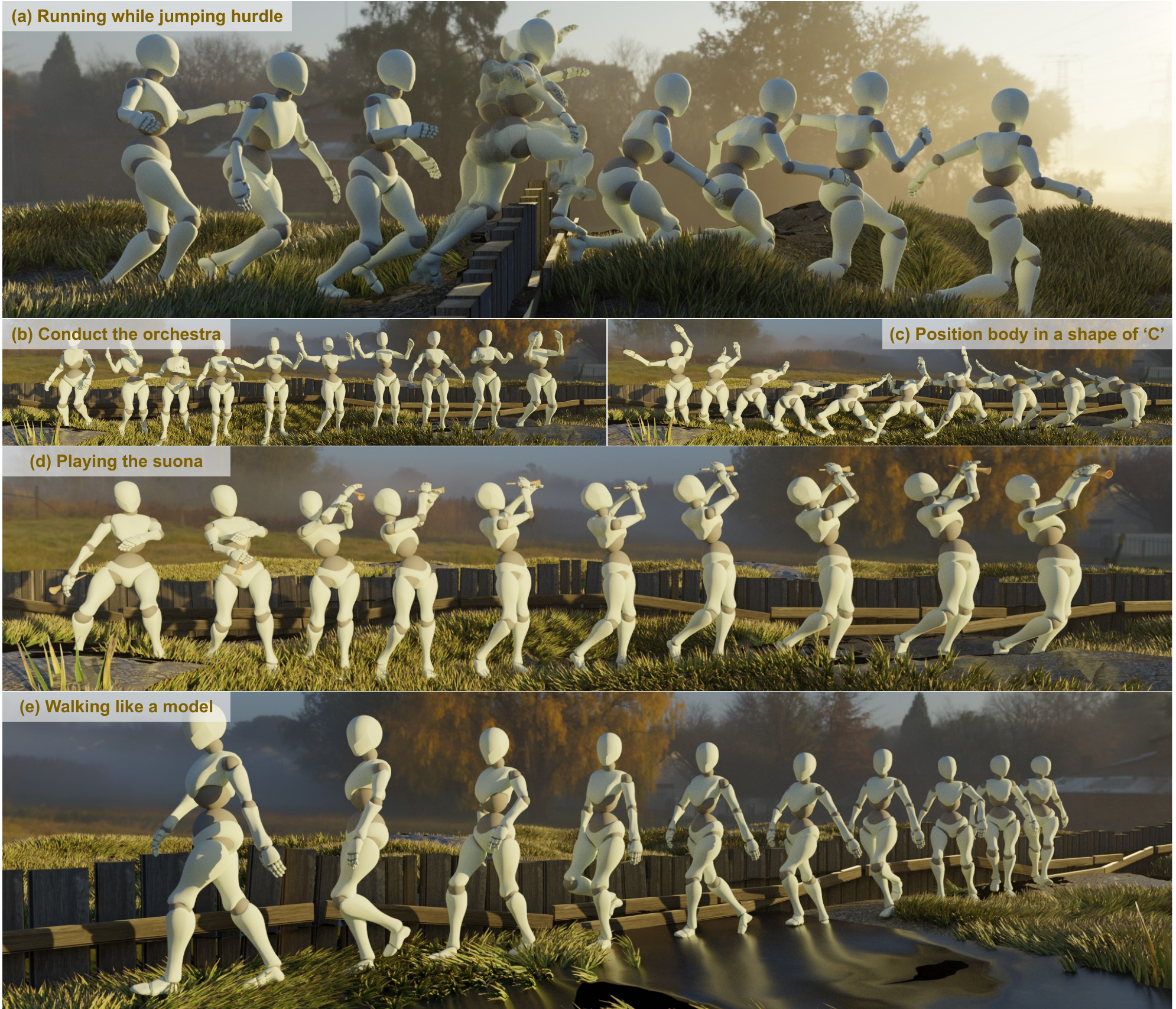}
    \caption{\textbf{Open-vocabulary humanoid skill synthesis using \model.} Our framework successfully generates physically plausible motions for diverse, previously \textbf{unseen} instructions. Each row demonstrates a sequence of frames depicting a different skill: (a) ``running while jumping hurdle'' showcases dynamic locomotion with obstacle navigation, (b) ``conduct the orchestra'' demonstrates expressive arm gestures, (c) ``position body in a shape of `C''' illustrates the understanding of abstract concepts, (d) ``playing the suona'' captures realistic hand positioning, and (e) ``walking like a model'' exhibits stylized locomotion with characteristic posture and gait patterns.}
    \label{fig:controller}
\end{figure*}

\subsection{Open-Vocabulary Humanoid Skill Acquisition}\label{exp:pre}

Generating open-vocabulary physical skills for humanoids presents significant challenges in motion synthesis. To fully leverage \model's capabilities, we employ a hierarchical control approach using a pre-trained low-level controller from CALM~\cite{tessler2023calm}, defined as $a = \pi_L(z, s)$, which generates actions $a$ based on the current state $s$ and a latent code $z$. We then train a high-level policy $z = \pi_H(s; I)$ that optimizes the \model reward for instruction $I$. This approach enables complex movements through instruction-guided coordination of motion primitives.

We conduct both qualitative and quantitative evaluations using five representative open-vocabulary instructions. For qualitative assessment, the resulting motions are visualized in \cref{fig:controller}, with additional sequences available on our project website. These demonstrations illustrate our method's ability to generate naturalistic motions that accurately reflect diverse textual instructions.

For quantitative assessment, we employ computational metrics and human judgment. The computational metrics include \textbf{Smoothness} (average change in acceleration across joints~\cite{zhang2021learning}) and \textbf{CLIP similarity} (cosine similarity between rendered frames and instruction text). For human evaluation, 30 participants rated each motion sequence on three dimensions (0-10 scale): \textbf{Task completion} (how effectively the motion fulfills the instruction), \textbf{Motion naturalness} (biological plausibility), and \textbf{Physics} (adherence to physical constraints). Participants underwent calibration with reference examples to ensure rating consistency.

We benchmark \model against state-of-the-art approaches in different paradigms: (i) \textbf{Motion retrieval}: TMR~\cite{petrovich2023tmr} uses extensive motion data sets, establishing a standard for naturalness. (ii) \textbf{Data-driven generation}: MotionGPT~\cite{jiang2024motiongpt} and MoMask~\cite{guo2024momask} represent cutting-edge approaches trained on text-motion paired datasets. (iii) \textbf{Zero-shot skill learning}: AnySkill~\cite{cui2024anyskill} and AvatarCLIP~\cite{hong2022avatarclip} generate motions from unseen text using \ac{vlm} supervision.

\cref{tab:comparison} reveals several insights: (i) TMR achieves high naturalness but lower task completion, indicating our instructions extend beyond existing text-motion dataset distributions. (ii) MoMask and MotionGPT demonstrate strong naturalness but lower task completion, reinforcing the limitations of purely data-driven approaches. (iii) AnySkill, relying exclusively on \ac{vlm}-based rewards, achieves the highest task completion among baselines but exhibits reduced naturalness, highlighting the inadequacy of single-modal rewards. (iv) \model outperforms all baselines in task completion while maintaining naturalness and smoothness, validating our multi-modal reward framework.

Notably, our evaluation presents a more stringent challenge as \model operates within physical simulation constraints, while most comparison methods (except AnySkill) produce motions without such constraints.

\begin{figure}[t!]
    \centering	
    \includegraphics[width=\linewidth]{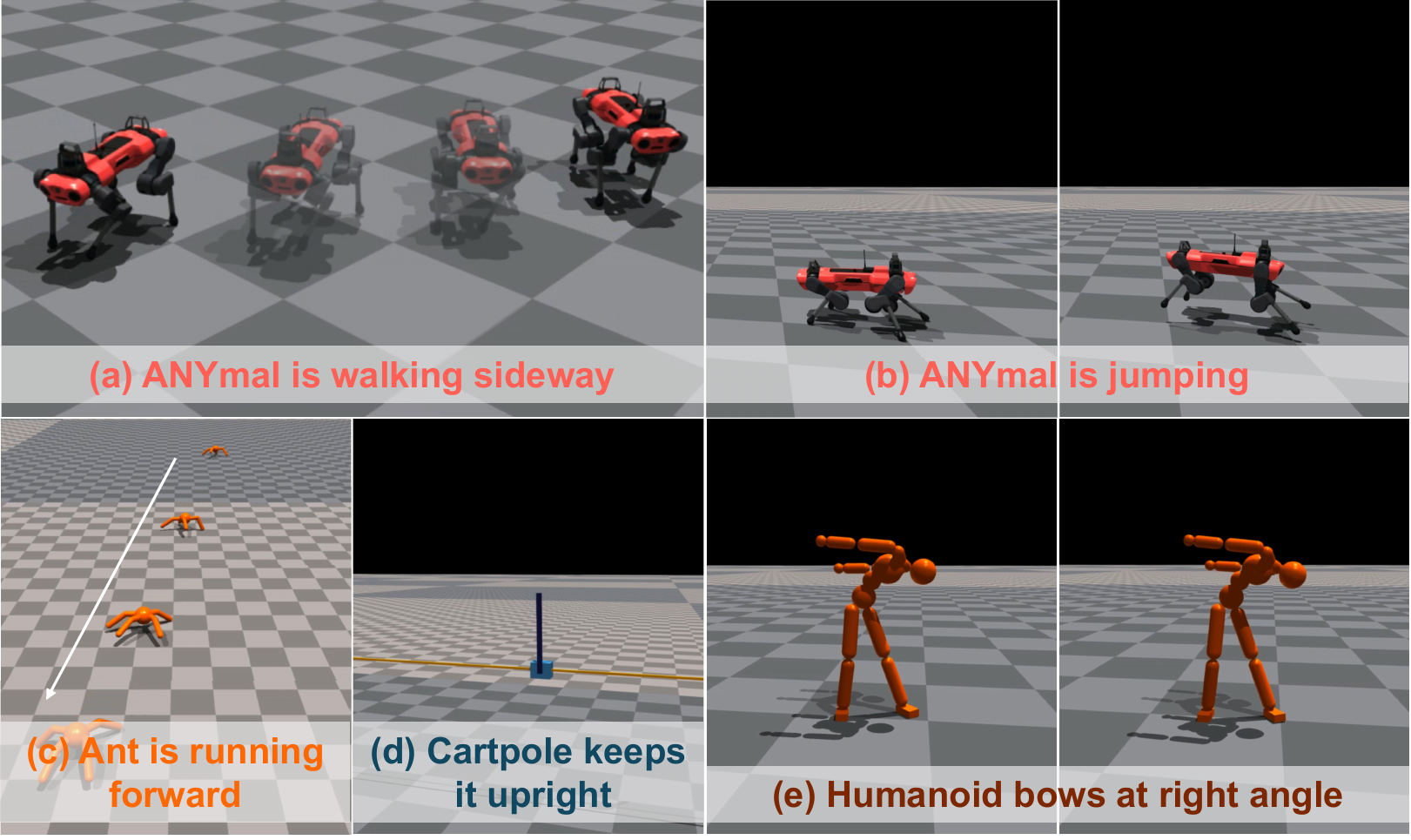}
    \caption{\textbf{Qualitative results of \model on standard \acs{rl} benchmarks, requiring to learn diverse control policies across different embodiments and task complexities.} Tasks shown include: ANYmal quadruped (a) moving sideways and (b) jumping, (c) Ant locomotion, (d) Cartpole balancing, (e) and Humanoid performing a bowing motion at a specific angle. Each task presents distinct control challenges requiring different physical capabilities. Additional benchmarks and videos are available on the project website.}
    \label{fig:RLbench}
\end{figure}

\begin{table}[b!]
    \centering
    \small
    \caption{\textbf{Quantitative evaluation on standard \acs{rl} benchmarks.} We compare \model against direct reward optimization across different agents and tasks. C represents the human-evaluated completion score, and RD (Reward Distance) learning efficiency. \model outperforms direct optimization in three of four tasks.}
    \label{tab:RLbench}
    \setlength{\tabcolsep}{3pt}
    \resizebox{\columnwidth}{!}{%
        \begin{tabular}{cccccc}
            \toprule
            \textbf{Agent} & \textbf{Task} & \textbf{C} $\uparrow$ & \textbf{RD (direct) $\downarrow$} & \textbf{RD (\model) $\downarrow$} \\
            \midrule
            Ant       & running forward & 8.897 & 103.81 & 92.45 \\
            Cartpole  & keep upright    & 8.862 & 20.72 & 20.04 \\
            ANYmal    & jump up         & 7.241 & 57.29 & 60.84 \\
            ANYmal    & walk sideways   & 8.414 & 39.71 & 37.51 \\
            \bottomrule
        \end{tabular}%
    }%
\end{table}

\subsection{Standard \texorpdfstring{\ac{rl}}{} Benchmarks}\label{exp:rl}

To establish \model's generalizability beyond humanoid control, we evaluate its performance across five diverse \ac{rl} tasks representing varying embodiments and complexity: Cartpole balancing (low-dimensional control), Ant walking (locomotion), ANYmal moving sideways (quadrupedal movement), ANYmal jumping (dynamic maneuver), and Humanoid bowing (articulated coordination), all illustrated in \cref{fig:RLbench}. For all experiments, we use three-layer MLP policies trained with PPO in IsaacGym.

Our evaluation combines human assessment with quantitative metrics. Following the protocol from \cref{exp:pre}, human evaluators rated task completion across all embodiments, consistently confirming \model's effectiveness. For quantitative comparison, we measure learning efficiency using the reward distance metric from \cite{rocamonde2024vlmrm}, capturing both maximum performance and convergence speed (see \cref{supp:exp:rlbench}).

Compared to a baseline that directly optimizes expert rewards, \model demonstrates superior training efficiency in three of four tasks, as shown in \cref{tab:RLbench}. The exception is the jumping task, where the baseline's advantage stems from directly optimizing the evaluation metric itself---a favorable scenario given this task's multiple success criteria.

A particularly revealing comparison is with \ac{vlm}-RM, which enhances agent textures specifically to improve \ac{vlm}-based reward effectiveness. As \cref{tab:RLhuman} demonstrates, our method outperforms \ac{vlm}-RM while using unmodified textures and only requiring 11.4\% of their training time. This efficiency differential highlights the advantage of our \ac{llm}-based reward formulation over approaches that rely on visual enhancements to the reward model.

\begin{table}[t!]
    \centering
    \small
    \caption{\textbf{Comparison with VLM-RM on humanoid bowing task.} T represents training time in minutes, and C human-evaluated completion score. \model outperforms \acs{vlm}-RM while requiring only 11.4\% of the training time.}
    \label{tab:RLhuman}
    \setlength{\tabcolsep}{3pt}
    \begin{tabular}{lllccc}
        \toprule
        \textbf{}     & \textbf{Task}  & \textbf{Textures} & \textbf{T}$\downarrow$ & \textbf{C}$\uparrow$  \\
        \midrule
        \ac{vlm}-RM ~\cite{rocamonde2024vlmrm}      &  Bow  &   Original    &   411min  &   1.655   \\
        \ac{vlm}-RM ~\cite{rocamonde2024vlmrm}      &  Bow  &   Improved    &   411min   &  3.483  \\ \midrule
        \textbf{\model}  &  Bow  &   Original    &   \textbf{47min}   &  \textbf{6.276}    \\
        \bottomrule
    \end{tabular}%
\end{table}

These results collectively demonstrate \model's robust generalizability across diverse embodied control tasks---from simple to complex embodiments and from standard locomotion to specialized movements. The consistent performance advantages suggest that \model's multi-modal reward structure effectively captures the essential characteristics of diverse tasks while providing smoother learning gradients than specialized reward engineering approaches.

\begin{table}[b!]
    \centering
    \small
    \caption{\textbf{Quantitative ablation study results.} We evaluate seven ablative model configurations across multiple metrics: completion quality (C), motion naturalness (N), movement smoothness (S), CLIP similarity (CLIP\_S), and training time (T).}
    \label{tab:ablation}
    \setlength{\tabcolsep}{3pt}
    \resizebox{\columnwidth}{!}{%
        \begin{tabular}{lccccc}
            \toprule
            \textbf{Ablations} & \textbf{C} $\uparrow$ & \textbf{N}$\uparrow$ & \textbf{S}$\downarrow$  & \textbf{CLIP\_S}$\uparrow$ & \textbf{T}$\downarrow$ \\
            \midrule
            \acs{vlm} only~\cite{cui2024anyskill}  &   6.108    &   5.938   &   0.486  & 23.925 &  59min \\
            \acs{vlm} + \acs{llm}    &   6.954   &   6.292   &   \textbf{0.457}  & 24.088     &  47min \\
            \ac{llm} only w/o \acs{rdp}~\cite{ma2024eureka}&  6.622 &  5.907 &  0.582  &  22.977 &  20min \\                \acs{llm} only with \acs{rdp} &   6.650   &  6.177   &  0.599   & 24.099    &  19min \\
            \posetoclip only  &   6.785    &    5.954   &  0.553   & 24.547    &  16min \\ 
            \posetoclip + \acs{llm}  w/o \acs{rdp} & 7.269   &  6.738  &  0.475 & 25.331  &  \textbf{7min} \\ 
            \posetoclip + \acs{llm}  &   \textbf{7.924}    &   \textbf{6.793}    &   0.488  & \textbf{28.998}   &  \textbf{7min} \\
            \bottomrule
        \end{tabular}%
    }%
\end{table}

\subsection{Ablations}\label{sec:ablation}

To analyze the contribution of each component in our approach, we conduct a comprehensive ablation study with seven model configurations, each isolating or combining different aspects of our full system: (i) \textbf{\ac{vlm} only} uses CLIP features extracted directly from simulation environment images, resembling the AnySkill~\cite{cui2024anyskill} baseline. (ii) \textbf{\ac{llm} only w/o \ac{rdp}} employs the \ac{llm}-generated reward function without dynamic refinement, similar to Eureka~\cite{ma2024eureka}. (iii) \textbf{\ac{llm} only (with \ac{rdp})} extends this by periodically regenerating the reward function based on fitness scores. (iv) \textbf{\ac{vlm} + \ac{llm}} combines both reward modalities with \ac{rdp}-based resampling of \ac{llm} rewards. (v) \textbf{\posetoclip only} substitutes standard CLIP features with our \posetoclip features in the \ac{vlm}-only setting. (vi) \textbf{\posetoclip + \ac{llm} w/o \ac{rdp}} combines \posetoclip with a static \ac{llm}-based reward generated once at training initialization. (vii) \textbf{\posetoclip + \ac{llm}} is our full model.

\begin{figure}[t!]
    \centering
    \includegraphics[width=\linewidth]{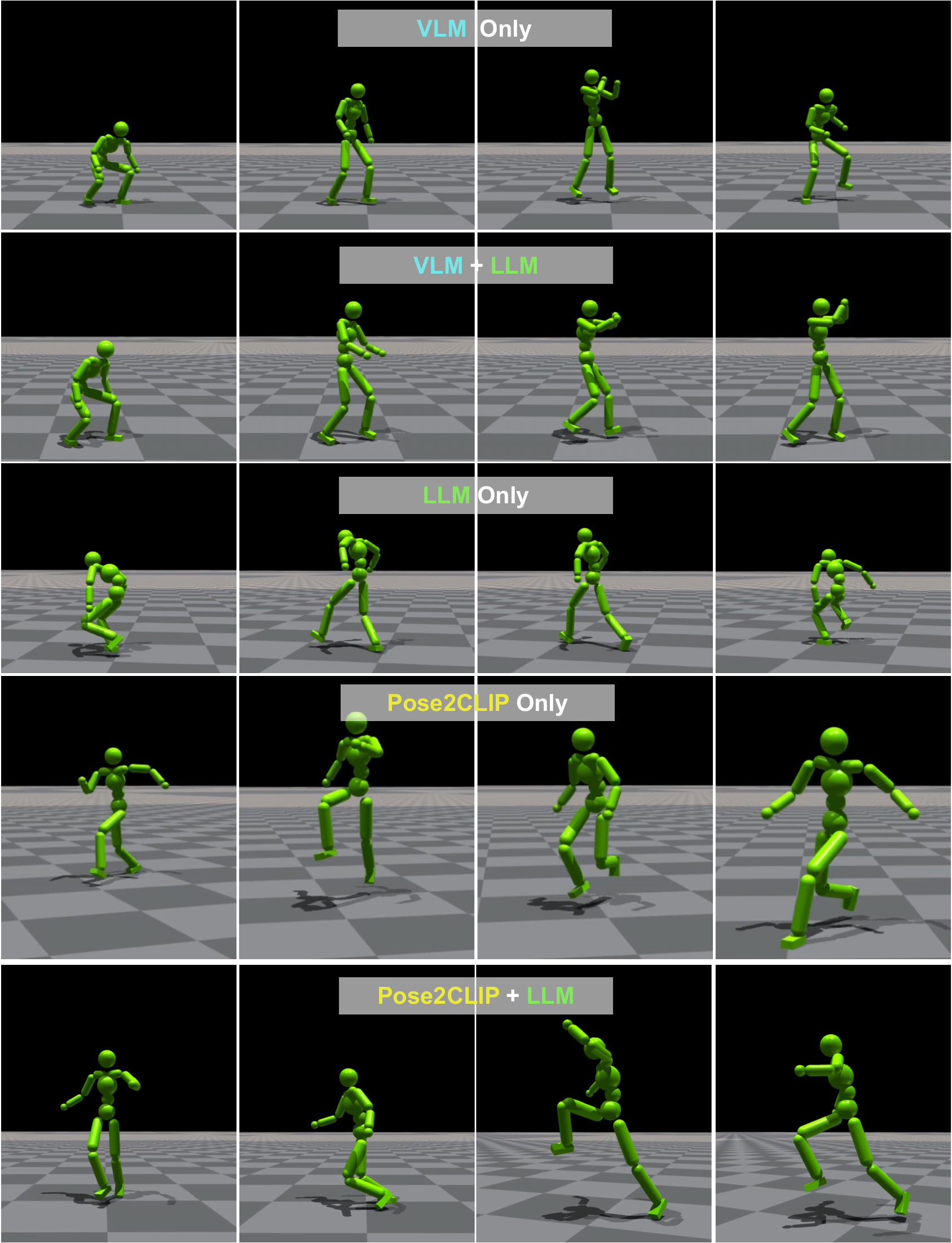}
    \caption{\textbf{Qualitative ablation study results.} We compare motion sequences generated by different model configurations for the instruction ``\textit{running while jumping hurdles}.'' The visualization reveals clear performance differences: \acs{vlm}-Only and \acs{llm}-Only approaches produce running motions with minimal vertical displacement, while \posetoclip configurations show distinct jumping actions with proper elevation and form. The full \posetoclip + \acs{llm} model generates the most natural hurdle-jumping motion, with appropriate preparation, elevation, and landing phases.}
    \label{fig:ablation}
\end{figure}

Qualitative results in \cref{fig:ablation} and quantitative evaluations in \cref{tab:ablation} reveal three key insights. (i) \posetoclip feature extraction improves task completion and reduces convergence time compared to direct CLIP features, underscoring the importance of high-quality visual representations for physically embodied tasks. (ii) \ac{llm}-only and \posetoclip-only configurations show complementary strengths---the former producing more natural motions, while the latter better adheres to instructional details---validating our hypothesis that these reward modalities capture different yet essential aspects of performance. (iii) Our full model consistently outperforms all variants across metrics, confirming the synergistic interaction between components and demonstrating that each element effectively addresses limitations of the others.

\begin{figure}[t!]
    \small
    \centering
    \begin{tabular}{@{}c@{\hspace{0cm}}c@{\hspace{0cm}}c@{}}
        \includegraphics[height=3.7cm]{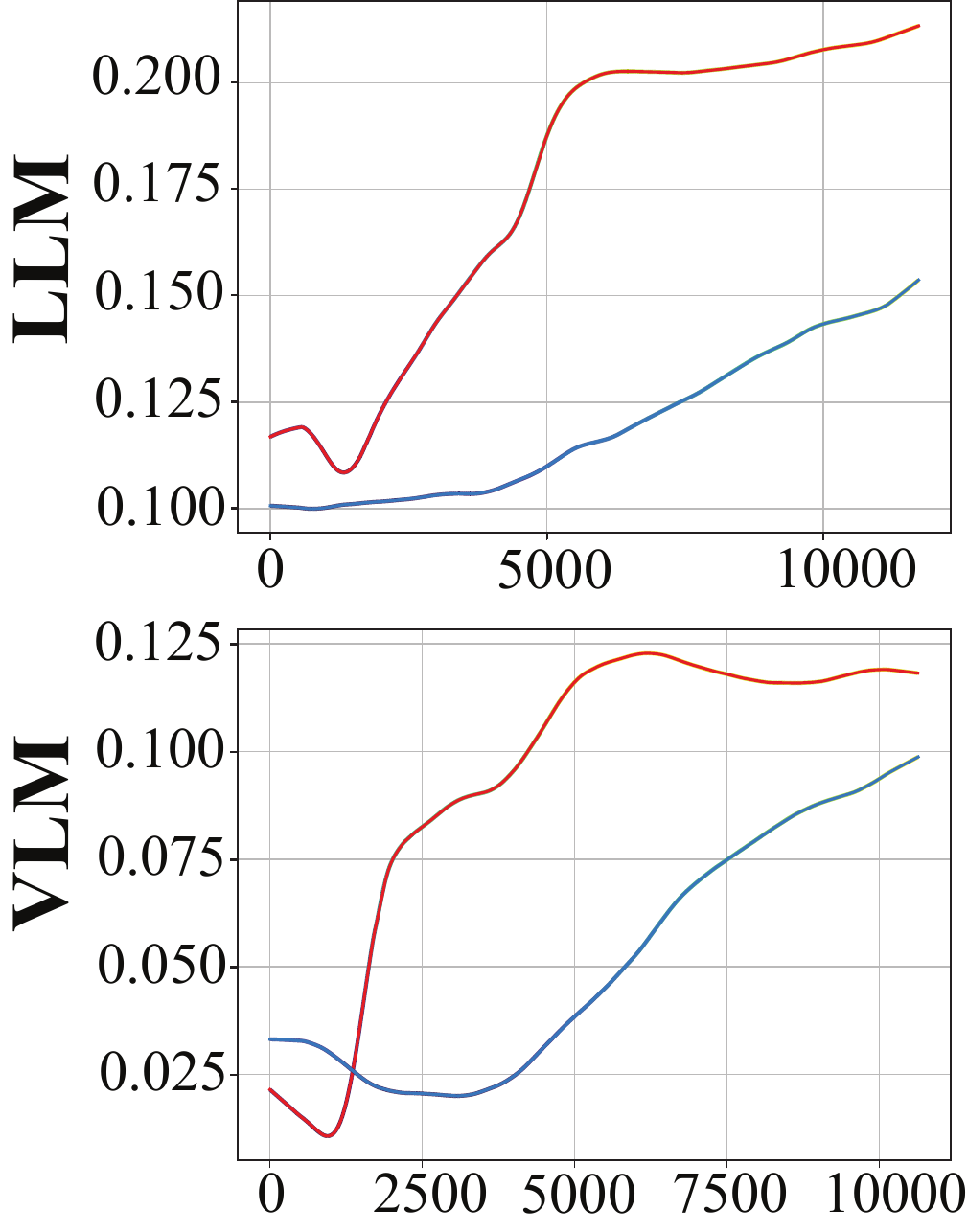} &
        \includegraphics[height=3.7cm]{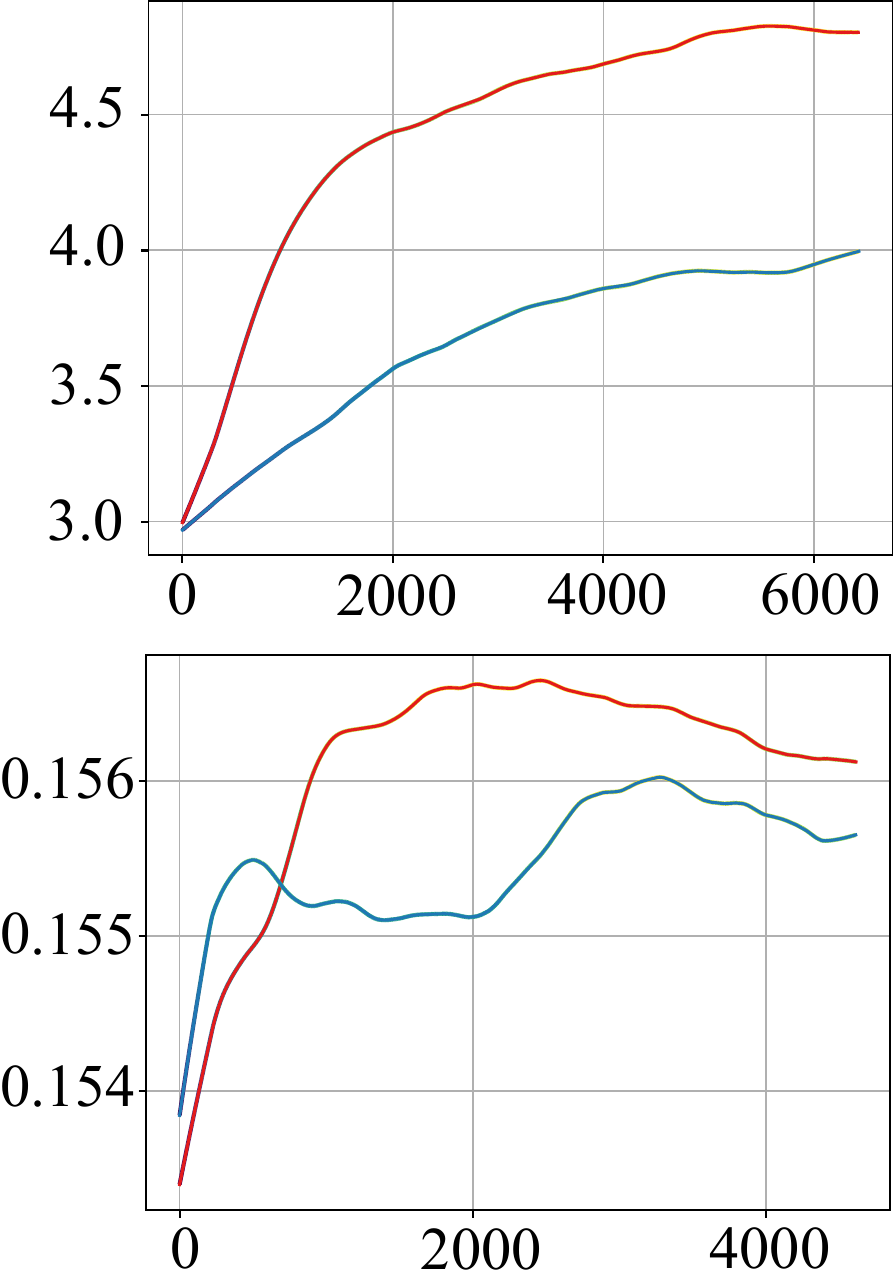} &
        \includegraphics[height=3.7cm]{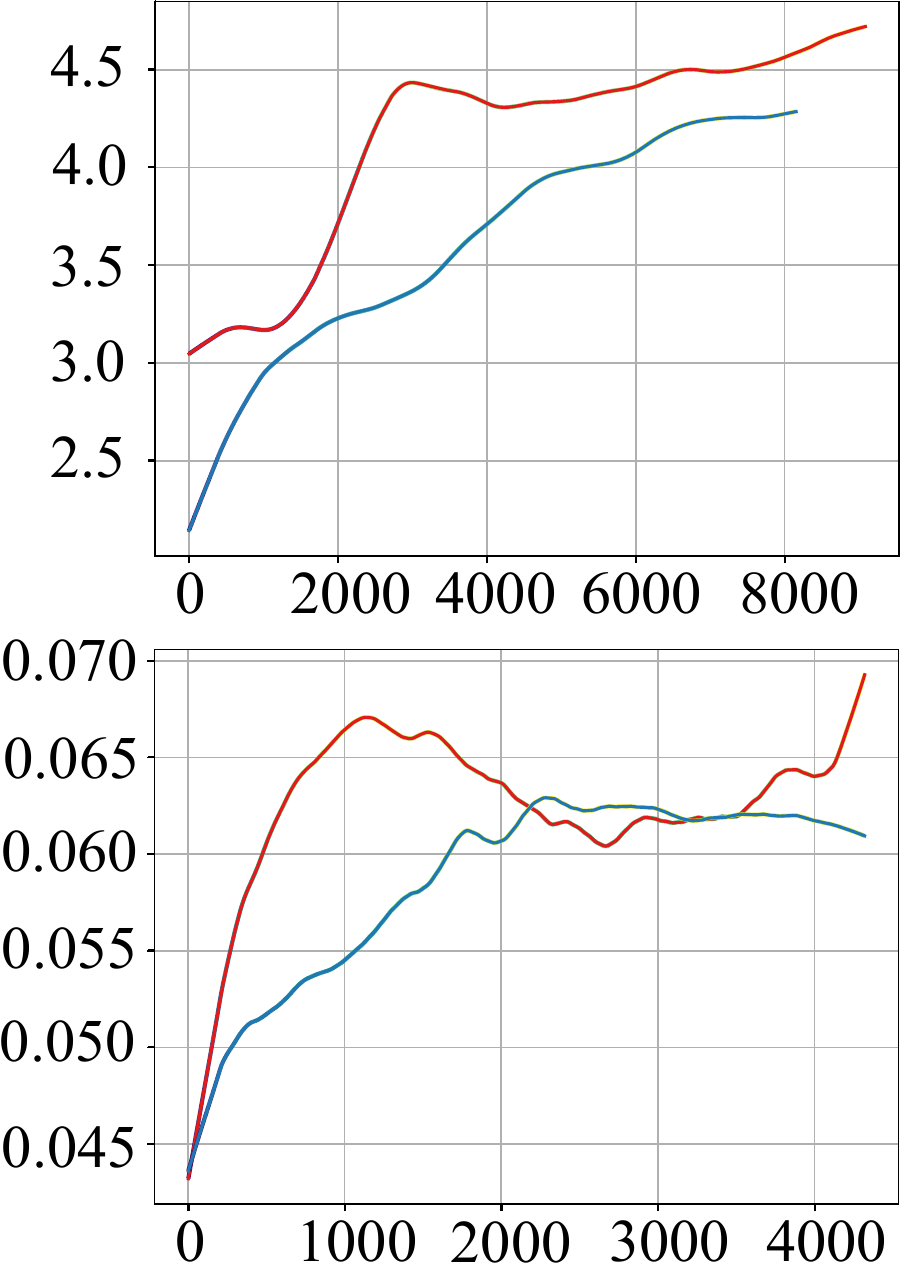} \\
        (a) & (b) & (c) \\
    \end{tabular}
    \caption{\textbf{Reward convergence comparison across tasks.} We compare reward trajectories between our full model (red) and single-reward baselines (blue) across three different tasks. Tasks: (a) arms folded over check; (b) position body in a shape of `C'; and (c) playing the suona.}
    \label{fig:abla:crv}
\end{figure}

The reward trajectories in \cref{fig:abla:crv} further support these findings by comparing our full model (red) against single-reward baselines (blue) across three tasks. The top row displays the \ac{llm} reward component, while the bottom row shows the \ac{vlm} component, with each baseline trained using only its corresponding reward type. These trajectories reveal a consistent pattern: the integration of multi-modal rewards accelerates the convergence of both reward functions compared to their uni-modal counterparts. This acceleration provides compelling evidence that \ac{llm}-based and \ac{vlm}-based rewards offer complementary guidance signals, enabling more efficient skill acquisition.

\section{Conclusion}

We present \model, a generalized framework for learning open-vocabulary physical skills that integrates \ac{vlm}-based reward computation, \ac{llm}-based reward generation, pose-to-semantic feature mapping, and dynamic reward programming. Our approach demonstrates significant advantages, improving task completion by 25.7\% while converging 8.4× faster than baselines, with comparable motion quality to specialized generative models. Experiments across standard \ac{rl} benchmarks confirm our method's generalizability, while ablation studies validate that our multimodal approach provides complementary guidance that accelerates skill acquisition. These results establish \model as an efficient framework for open-vocabulary skill learning in robotic control and motion synthesis.

\paragraph{Acknowledgment}

This work is supported in part by the National Natural Science Foundation of China (62376009), the Beijing Nova Program, the State Key Lab of General AI at Peking University, the PKU-BingJi Joint Laboratory for Artificial Intelligence, and the National Comprehensive Experimental Base for Governance of Intelligent Society, Wuhan East Lake High-Tech Development Zone.
\clearpage
{
    \small
    \setstretch{0.97}
    \bibliographystyle{ieeenat_fullname}
    \bibliography{reference_header,reference}
}

\clearpage
\appendix
\renewcommand\thefigure{A\arabic{figure}}
\setcounter{figure}{0}
\renewcommand\thetable{A\arabic{table}}
\setcounter{table}{0}
\renewcommand\theequation{A\arabic{equation}}
\setcounter{equation}{0}
\pagenumbering{arabic}% resets `page` counter to 1
\renewcommand*{\thepage}{A\arabic{page}}
\setcounter{footnote}{0}

\begingroup
\let\clearpage\relax 
\onecolumn

\section{Additional Experimental Results}

\subsection{Open-Vocabulary Humanoid Skill Acquisition}

While our main paper presents compelling quantitative evidence of \model's effectiveness, the qualitative differences become even more striking when visualized. In \cref{supp:fig:baseline1,supp:fig:baseline2}, we showcase these differences across nine distinct approaches: (i) MoMask~\cite{guo2024momask}, (ii) MotionGPT~\cite{jiang2023motiongpt}, (iii) TMR~\cite{petrovich2023tmr}, (iv) AvatarCLIP~\cite{hong2022avatarclip}, (v) AnySkill~\cite{cui2024anyskill}, (vi) \ac{vlm}+\ac{llm}, (vii) \ac{llm} only, (viii) \posetoclip only, and (ix) our full \model{} framework.

To thoroughly evaluate generalization capabilities, we challenged each method with diverse open-vocabulary instructions: (i) ``playing the suona'', (ii) ``running while jumping hurdle'', (iii) ``conduct the orchestra'', (iv) ``walking like a model'', and (v) ``position body in a shape of `C'''. The results reveal patterns across model categories. TMR, despite its sophistication, consistently fails to generate appropriate responses---a clear indication that these open-vocabulary prompts lie beyond the distribution of current text-motion datasets. MoMask demonstrates competence with familiar actions like ``jumping hurdles'' but falters when confronted with more novel or culturally specific instructions such as ``playing the suona.'' Similarly, MotionGPT tends to produce conservative motions that only partially capture the intended behaviors, often defaulting to upright, small-amplitude movements that lack expressivity. AvatarCLIP likewise demonstrates limitations in translating textual instructions into fluid, meaningful motions.

Pushing the boundaries further, we designed an additional challenge set of instructions carefully crafted to satisfy two critical criteria:   
\begin{itemize}[leftmargin=*,nolistsep,noitemsep]
    \item They emerge from an \ac{llm}'s understanding of natural human movement patterns.
    \item They verifiably exist outside the domain of any open-source human text-motion dataset.
\end{itemize}

This design principle ensures that success cannot be attributed to mere retrieval of existing motion data---a true test of generative understanding. The challenge set includes diverse instructions such as ``jump rope,'' ``walking while sipping water,'' ``swim with two arms,'' ``throw a ball, one hand scratch forward,'' ``hurrah with two arms,'' and ``jump in place.'' These prompts require ian ntegrated understanding of physics, biomechanics, and semantic intent.

To capture the multidimensional nature of motion quality, our human evaluation framework employs three complementary metrics, each rated on a 0--10 scale:
\begin{itemize}[leftmargin=*,nolistsep,noitemsep]
    \item \textbf{Task completion:} Measures semantic fidelity---how faithfully the motion embodies the instruction's intent. Higher scores reflect more accurate realization of the specified action.
    \item \textbf{Motion naturalness:} Evaluates kinematic plausibility---the smoothness and continuity of movement patterns. This metric penalizes jarring transitions, unnatural accelerations, or anatomically implausible configurations.
    \item \textbf{Physics:} Assesses physical realism---how well the motion respects fundamental physical constraints. Higher scores indicate fewer artifacts like floating, ground penetration, or impossible joint relationships.
\end{itemize}

Together, these metrics provide a holistic assessment framework that aligns with human perception of motion quality. The framework reveals \model's ability to generate motions that not only accomplish the specified task but do so with natural, physically plausible movements---a significant advancement over existing approaches. For a more visceral understanding of these qualitative differences, we encourage readers to explore the additional videos and interactive visualizations available on our project website.

\begin{figure}[t!]
    \centering
    \includegraphics[width=\linewidth]{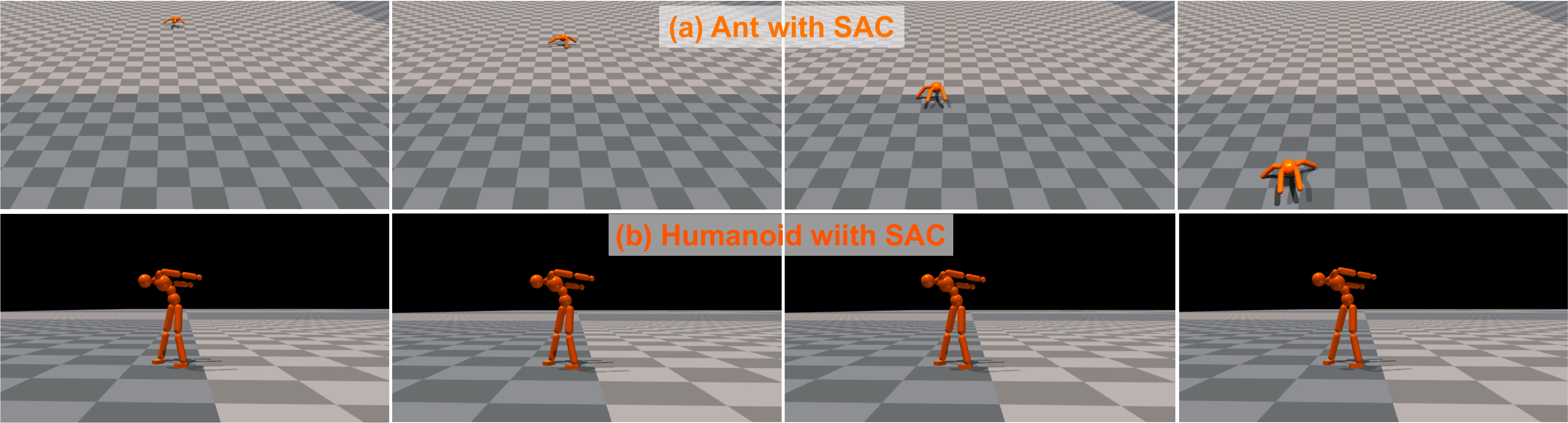}
    \caption{\textbf{Training standard \acs{rl} environments with \model{} using SAC.} (a) The Ant agent executing ``running forward'' behavior---note the consistent forward progress across frames. (b) The Humanoid agent performing a ``bow at a right angle''---observe how the agent maintains balance while achieving the specified angle. These results demonstrate that \model{} effectively guides agent behavior regardless of whether PPO or SAC is used as the underlying \acs{rl} algorithm.}
    \label{supp:fig:rl}
\end{figure}

\subsection{Standard \texorpdfstring{\ac{rl}}{} Benchmarks}

Beyond the standard \ac{rl} benchmark results presented in our main paper, we conducted additional experiments to evaluate \model's compatibility with alternative \ac{rl} algorithms. While our primary results utilize Proximal Policy Optimization (PPO), we also tested our approach with Soft Actor-Critic (SAC), a different \ac{rl} algorithm with distinct optimization characteristics.

For this comparison, we focused on two challenging environments from our benchmark suite: Humanoid and Ant. These environments feature complex dynamics and high-dimensional action spaces that provide a rigorous test of our framework's capabilities. To ensure direct comparability, we maintained the same text instructions used in our PPO experiments: ``Humanoid bows at a right angle'' for the Humanoid environment and ``Ant is running forward'' for the Ant environment.

The results, visualized in \cref{supp:fig:rl}, demonstrate that \model{} successfully enables agents to perform the actions specified by the text instructions across both \ac{rl} algorithms. This consistency across different optimization methods suggests that our multimodal reward framework provides effective guidance regardless of the underlying algorithm choice.

These findings complement our main results by showing that \model's approach to open-vocabulary skill acquisition generalizes beyond a single \ac{rl} algorithm, further supporting its utility as a flexible framework for teaching diverse behaviors to simulated agents.

\section{\ac{llm}-based Reward}\label{supp:method:llm}

This section provides a comprehensive overview of our approach to designing and implementing \ac{llm}-based reward functions. We describe both the carefully engineered prompts that elicit effective reward functions and the resulting outputs from various \ac{llm} models. Our prompt engineering process addresses several critical requirements for embodied reinforcement learning with natural language guidance.

\subsection{Prompt Input}

\begin{breakbox}
You are a reward engineer trying to write reward functions to solve \ac{rl} tasks as effectively as possible. 

Your goal is to write a reward function for the environment that will help a humanoid character learn the task described in the text. The humanoid character is a physics-based character with 15 joints. There are other components in the model that keep the humanoid upright and moving like a human being, so your job is only to write a reward function that captures the essence of the described task. 

The joint order for the humanoid is as follows:
\begin{lstlisting}
SMPL_BONE_ORDER_NAMES = [
    "pelvis",
    "torso",
    "head",
    "right_upper_arm",
    "right_lower_arm",
    "right_hand",
    "left_upper_arm",
    "left_lower_arm",
    "left_hand",
    "right_thigh",
    "right_shin",
    "right_foot",
    "left_thigh",
    "left_shin",
    "left_foot"
]
\end{lstlisting}

In the simulator, we define the z as the up-axis.

Your reward function should use useful variables from the environment as inputs. As an example, the reward function signature can be: 
\begin{lstlisting}
@torch.jit.script
def compute_llm_reward():
    body_pos = infos["state_embeds"][:, :21, :3]  # [num, 21, 3]
    body_rot = infos["state_embeds"][:, :21, 3:7]  # [num, 21, 4]
    body_vel = infos["state_embeds"][:, :21, 7:10]  # [num, 21, 3]
    body_ang_vel = infos["state_embeds"][:, :21, 10:13]  # [num, 21, 3]
    ...
    return reward, {}
\end{lstlisting}

You can parse each joint's position, rotation, velocity, and angular velocity from the tensors as follows:

\begin{lstlisting}
pelvis_pos, pelvis_rot, pelvis_vel, pelvis_ang_vel = body_pos[:, 0, :], body_rot[:, 0, :], body_vel[:, 0, :], body_ang_vel[:, 0, :]
torso_pos, torso_rot, torso_vel, torso_ang_vel = body_pos[:, 1, :], body_rot[:, 1, :], body_vel[:, 1, :], body_ang_vel[:, 1, :]
head_pos, head_rot, head_vel, head_ang_vel = body_pos[:, 2, :], body_rot[:, 2, :], body_vel[:, 2, :], body_ang_vel[:, 2, :]
...
\end{lstlisting}

the order is the same as the \texttt{SMPL\_BONE\_ORDER\_NAMES}.

Since the reward function will be decorated with $@$torch.jit.script, please make sure that the code is compatible with TorchScript (e.g., use torch tensor instead of numpy array). 

Make sure any new tensor or variable you introduce is on the same device as the input tensors. The output of the reward function should consist of two items:
\begin{itemize}
    \item the total reward,
    \item a dictionary of each individual reward component.
\end{itemize}

The code output should be formatted as a python code string: \texttt{``` python ... ```}.

Some helpful tips for writing the reward function code:
\begin{enumerate}
    \item You may find it helpful to normalize the reward to a fixed range by applying transformations like torch.exp to the overall reward or its components
    \item If you choose to transform a reward component, then you must also introduce a temperature parameter inside the transformation function; this parameter must be a named variable in the reward function and it must not be an input variable. Each transformed reward component should have its own temperature variable
    \item Make sure the type of each input variable is correctly specified; a float input variable should not be specified as torch.Tensor
    \item Most importantly, the reward code\'s input variables must contain only attributes of the provided environment class definition (namely, variables that have prefix self.). Under no circumstance can you introduce new input variables.
\end{enumerate}
   
Write a reward function for the following task: \textbf{{open-vocabulary instruction}}
\end{breakbox}

\subsection{\ac{llm} Reward Generated by Different Baselines}

To determine the most effective \ac{llm} for reward generation, we conducted a systematic comparison using multiple language instructions. Here, we examine the outputs for a representative example: ``boxing with two arms.'' We evaluated three prominent \ac{llm} models by analyzing their generated reward functions for correctness, physical plausibility, and alignment with the intended behavior. The evaluation extended beyond this single example to include more than five diverse text prompts that tested different aspects of motion understanding.

Our analysis revealed that while all tested models produced semantically reasonable reward structures, they differed significantly in their handling of physical constraints and motion dynamics. Specifically, \textit{GPT-4o} demonstrated superior performance by consistently avoiding physically impossible reward terms or biomechanically unrealistic motion specifications.

Based on these comparative results, we selected \textit{GPT-o1-preview} as our primary \ac{llm}-based reward generator for all experiments reported in this paper. This choice ensures that our baseline comparisons reflect the strongest possible implementation of pure \ac{llm}-based reward approaches.

\begin{breakbox1}
\begin{lstlisting} 
@torch.jit.script
def compute_llm_reward(self):
    body_pos = self.infos["state_embeds"][:, :15, :3]  # [num, 15, 3]
    body_rot = self.infos["state_embeds"][:, :15, 3:7]  # [num, 15, 4]
    body_vel = self.infos["state_embeds"][:, :15, 7:10]  # [num, 15, 3]
    body_ang_vel = self.infos["state_embeds"][:, :15, 10:13]  # [num, 15, 3]

    # Extract positions
    torso_pos = body_pos[:, 1, :]         # "torso" is at index 1
    left_hand_pos = body_pos[:, 8, :]     # "left_hand" is at index 8
    right_hand_pos = body_pos[:, 5, :]    # "right_hand" is at index 5

    # Compute distances to torso
    left_hand_to_torso_dist = torch.norm(left_hand_pos - torso_pos, dim=1)
    right_hand_to_torso_dist = torch.norm(right_hand_pos - torso_pos, dim=1)
    
    # Compute distance between hands
    hands_distance = torch.norm(left_hand_pos - right_hand_pos, dim=1)

    # Temperature parameters for transformations
    temp_hand_torso = 0.2
    temp_hands = 0.2

    # Compute reward components using exponential decay based on distances
    left_hand_reward = torch.exp(-left_hand_to_torso_dist / temp_hand_torso)
    right_hand_reward = torch.exp(-right_hand_to_torso_dist / temp_hand_torso)
    hands_proximity_reward = torch.exp(-hands_distance / temp_hands)

    # Total reward is the average of the three components
    llm_rewards = (left_hand_reward + right_hand_reward + hands_proximity_reward) / 3.0

    # Dictionary of individual reward components
    reward_components = {
        "left_hand_reward": left_hand_reward,
        "right_hand_reward": right_hand_reward,
        "hands_proximity_reward": hands_proximity_reward
    }

    return llm_rewards
    \end{lstlisting}
\end{breakbox1}

\begin{breakbox2}
To create a reward function for the humanoid to perform boxing, we need to focus on rewarding actions that resemble boxing movements. This involves encouraging the humanoid to use its arms (specifically the shoulders, elbows, and wrists) actively, maintain balance, and possibly target specific points in space to simulate punches.

Here's an example reward function:
\begin{lstlisting}
@torch.jit.script
def compute_boxing_reward(
    self
) -> Tuple[torch.Tensor, Dict[str, torch.Tensor]]:
    body_pos = self._rigid_body_pos  # [num, 15, 3]
    body_rot = self._rigid_body_rot  # [num, 15, 4]
    body_vel = self._rigid_body_vel  # [num, 15, 3]
    body_ang_vel = self._rigid_body_ang_vel  # [num, 15, 3]

    left_shoulder_vel = body_vel[:, 16, :]
    right_shoulder_vel = body_vel[:, 17, :]
    left_elbow_vel = body_vel[:, 18, :]
    right_elbow_vel = body_vel[:, 19, :]
    left_wrist_vel = body_vel[:, 20, :]
    right_wrist_vel = body_vel[:, 21, :]

    # Reward for arm movement (boxing punches)
    arm_movement_reward = (
        torch.norm(left_shoulder_vel, dim=1)
        + torch.norm(right_shoulder_vel, dim=1)
        + torch.norm(left_elbow_vel, dim=1)
        + torch.norm(right_elbow_vel, dim=1)
        + torch.norm(left_wrist_vel, dim=1)
        + torch.norm(right_wrist_vel, dim=1)
    )

    # Maintain balance by keeping pelvis stable
    pelvis_pos = body_pos[:, 0, :]
    pelvis_stability_reward = -torch.abs(pelvis_pos[:, 2] - 1.0)  # Assuming 1.0 is the desired pelvis height

    # Encourage facing forward by aligning the pelvis rotation with the forward direction
    forward_direction = torch.tensor([0.0, 0.0, 1.0], device=body_rot.device)
    pelvis_rot = body_rot[:, 0, :]
    facing_reward = torch.sum(pelvis_rot[:, :3] * forward_direction, dim=1)

    # Transform rewards with temperature parameters
    arm_movement_temp = 0.1
    stability_temp = 0.1
    facing_temp = 0.1

    arm_movement_reward = torch.exp(arm_movement_reward * arm_movement_temp)
    pelvis_stability_reward = torch.exp(pelvis_stability_reward * stability_temp)
    facing_reward = torch.exp(facing_reward * facing_temp)

    # Total reward
    total_reward = arm_movement_reward + pelvis_stability_reward + facing_reward

    reward_components = {
        "arm_movement_reward": arm_movement_reward,
        "pelvis_stability_reward": pelvis_stability_reward,
        "facing_reward": facing_reward,
    }

    return total_reward, reward_components
\end{lstlisting}
\end{breakbox2}

\begin{breakbox3}
\begin{lstlisting}
@torch.jit.script
def compute_llm_reward(
    body_pos: torch.Tensor,
    body_rot: torch.Tensor,
    body_vel: torch.Tensor,
    body_ang_vel: torch.Tensor,
) -> Tuple[torch.Tensor, Dict[str, torch.Tensor]]:
    """
    Reward function for making the humanoid box.

    Args:
        body_pos: Position of each body part.
        body_rot: Rotation of each body part.
        body_vel: Velocity of each body part.
        body_ang_vel: Angular velocity of each body part.

    Returns:
        A tuple containing the total reward and a dictionary of reward components.
    """
    device = body_pos.device

    # Extract left and right hand positions and velocities.
    left_hand_pos = body_pos[:, SMPL_BONE_ORDER_NAMES.index("left_hand"), :]
    right_hand_pos = body_pos[:, SMPL_BONE_ORDER_NAMES.index("right_hand"), :]
    left_hand_vel = body_vel[:, SMPL_BONE_ORDER_NAMES.index("left_hand"), :]
    right_hand_vel = body_vel[:, SMPL_BONE_ORDER_NAMES.index("right_hand"), :]

    # Reward for hands close together (forming a boxing stance).
    hand_dist = torch.norm(left_hand_pos - right_hand_pos, dim=-1)
    hands_together_reward = torch.exp(-hand_dist * 2.0)

    # Reward for hands at head level.
    head_pos = body_pos[:, SMPL_BONE_ORDER_NAMES.index("head"), :]
    left_hand_head_diff = torch.abs(left_hand_pos[:, 1] - head_pos[:, 1]) # y-axis difference
    right_hand_head_diff = torch.abs(right_hand_pos[:, 1] - head_pos[:, 1]) # y-axis difference
    hands_at_head_level_reward = torch.exp(-(left_hand_head_diff + right_hand_head_diff) * 2.0)


    # Reward for minimizing hand velocity (stable boxing stance).
    left_hand_vel_magnitude = torch.norm(left_hand_vel, dim=-1)
    right_hand_vel_magnitude = torch.norm(right_hand_vel, dim=-1)
    hands_velocity_reward = torch.exp(-(left_hand_vel_magnitude + right_hand_vel_magnitude) * 0.5)

    # Combine reward components.
    reward = hands_together_reward + hands_at_head_level_reward + hands_velocity_reward

    # Reward dictionary.
    reward_dict = {
        "hands_together_reward": hands_together_reward,
        "hands_at_head_level_reward": hands_at_head_level_reward,
        "hands_velocity_reward": hands_velocity_reward,
    }

    return reward, reward_dict
\end{lstlisting}
\end{breakbox3}

\section{Additional Details of Experiments}

\subsection{Ablative Analysis of RDP}

Drawing inspiration from Eureka~\cite{ma2024eureka}, we formulate \model as a \acf{rdp}~\cite{singh2009rewards} where a \ac{vlm} provides feedback on the quality of language-generated rewards. In our framework, a \ac{vlm}-based reward $R_V$ acts as an evaluator for the \ac{llm}-based reward $R_L$, establishing a mutually reinforcing relationship between the two reward components.

To implement this interaction effectively, we develop a dynamic reward regeneration mechanism. Specifically, we trigger the regeneration of the \ac{llm}-based reward $R_L$ when we detect a sustained decline in performance---defined as eight consecutive steps with decreasing average $R_V$ across all parallel environments, with the latest average falling below the threshold of 0.1. This rejection-based sampling approach serves as a critical quality control mechanism, preventing the optimization of $R_L$ from diverging toward behaviors that may be mathematically optimal but visually inconsistent with the specified instruction.

To evaluate the importance of this \ac{rdp} mechanism, \cref{supp:fig:rdp} presents systematic qualitative comparisons across five diverse text commands: (i) ``running while jumping hurdle'', (ii) ``playing the suona,'' (iii) ``walking like a model,'' (iv) ``position body in a shape of `C','' and (v) ``conduct the orchestra.'' Our project website provides additional examples for a comprehensive review.

The ablation results reveal two critical insights about the \ac{rdp} mechanism. First, eliminating the rewriting component significantly reduces the effectiveness of constraints for certain instructions, causing generated behaviors to deviate from the specified text requirements. Second, actions generated without \ac{rdp} frequently resemble those produced by the \posetoclip component in isolation, indicating insufficient integration of the \ac{llm}'s semantic understanding. This lack of synergistic improvement---where the combined reward fails to outperform its individual components---underscores the importance of the \ac{rdp} mechanism in \model{}'s design for achieving robust instruction-following behavior.

\begin{figure}[t!]
    \centering
    \includegraphics[width=.7\linewidth]{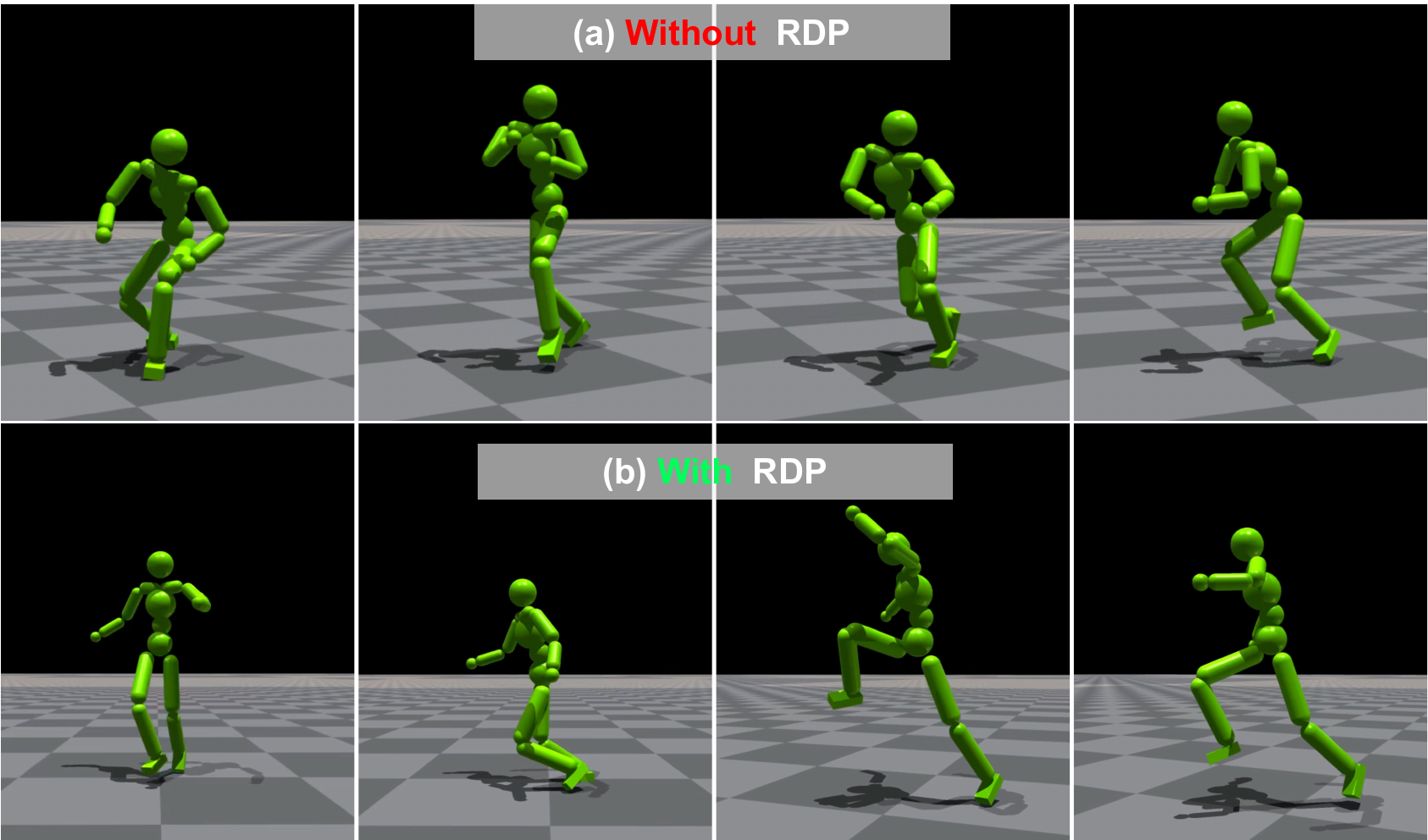}
    \caption{\textbf{Comparative performance with and without \acs{rdp}.} We illustrate the qualitative difference in motion execution for the instruction ``running while jumping hurdle.'' \textbf{(a) Without \acs{rdp}:} The agent exhibits limited vertical displacement and fails to perform a proper hurdle-clearing motion, instead executing a minimal hop that would be insufficient to clear an obstacle. Note the relatively flat trajectory and diminished preparation phase. \textbf{(b) With \acs{rdp}:} The agent demonstrates significantly improved biomechanics with proper jump preparation, extension during flight, and recovery phases. Observe the pronounced knee lift, extended flight phase, and appropriate body orientation---all critical elements of successful hurdle clearance. This comparison highlights how \acs{rdp} substantively improves motion quality by dynamically refining \acs{llm}-based rewards through \acs{vlm} feedback, resulting in semantically accurate and physically plausible behaviors.}
    \label{supp:fig:rdp}
\end{figure}

\subsection{Quantifying Policy Convergence with Expert Reward Distance}\label{supp:exp:rlbench}

To quantitatively evaluate policy learning efficiency across different reward formulations, we developed a standardized metric called reward distance, reported in \cref{tab:RLbench}. This metric measures how quickly a policy converges to optimal behavior according to task-specific expert criteria, regardless of the actual reward function used during training.

For each \ac{rl} benchmark, we first define an expert reward function that captures the fundamental objective of the task. For instance, in the \textit{Ant Running Forward} task, the expert reward is expressed as $r_\mathrm{expert} = v_y$, where $v_y$ represents the agent's forward velocity---a direct measure of how well the agent accomplishes the primary goal of forward movement.

Importantly, we track these expert reward values independently from whatever reward function is actually used during policy training. This allows us to fairly compare different reward formulations based on their effectiveness at accomplishing the core task objective. The reward distance is then computed as the area above the expert reward curve throughout training. Lower values indicate faster convergence to optimal behavior, providing a quantitative measure of training efficiency that enables direct comparison across different reward approaches.

\subsection{Human Evaluation of Motion Quality and Task Alignment}

To rigorously evaluate the perceptual quality of generated motions beyond computational metrics, we conducted a comprehensive user study involving 30 participants with diverse backgrounds and varying levels of familiarity with computer animation and motion synthesis.

We employed a within-subjects experimental design, where each participant evaluated multiple motion sequences generated by different methods. To ensure methodological rigor, we incorporated several controls: (i) repeated items to assess participant consistency, (ii) randomized presentation order to mitigate sequence effects, and (iii) balanced exposure to different motion types across participants to control for potential biases.

The study was implemented through a custom web-based platform that presented motion sequences with standardized viewing angles and playback speeds. To minimize evaluation fatigue and maintain data quality, the system assigned each participant a unique task sequence with optimized pacing and appropriate rest intervals. Participants rated motions on multiple dimensions including semantic accuracy, physical plausibility, and overall naturalness, providing a multifaceted assessment of motion quality that complements our quantitative metrics.

\section{\posetoclip: Architecture and Implementation}

\subsection{Dataset Construction and Processing}

The effectiveness of our \posetoclip model relies on its exposure to diverse and representative pose data. To achieve this, we constructed a comprehensive training dataset integrating multiple high-quality motion sources. Our primary data foundations include the SMPL model poses from the AMASS training split and the extensive Motion-X dataset, specifically incorporating IDEA400 and four additional specialized subsets (Animation, HuMMan, Kungfu, and Perform) that capture a wide range of human movements.

To enhance robustness to physical simulator artifacts, we implemented an iterative data enrichment strategy. Pose data generated during the reinforcement learning training process is automatically captured, re-oriented to canonical coordinates, and incorporated back into the model's training dataset. This approach creates a diverse learning environment that includes both naturalistic human movements from motion capture and poses that may deviate from typical human motion patterns, particularly those generated during early exploration phases of the reinforcement learning process.

To maintain computational efficiency and prevent overfitting to redundant examples, we applied a systematic downsampling procedure for deduplication, particularly for highly similar poses occurring in repetitive motions. The precise distribution of data sources and their relative contributions to our final training dataset are detailed in \cref{supp:tab:data}.

\begin{table}[t!]
    \centering
    \small
    \caption{\textbf{Dataset Composition for \posetoclip Training.} We detail the diverse sources and sampling strategies used to create \posetoclip training dataset. While AMASS and IDEA400 provide the foundation with high-quality motion capture data (together contributing over 36 hours), we incorporated additional Motion-X subsets at full resolution to capture specialized movement patterns. To enhance robustness to simulation-specific poses and reduce distributional shift during deployment, we augmented the dataset with frames sampled directly from our reinforcement learning training episodes. The varying sampling rates (1/5 for most sources) were implemented to balance computational efficiency with representation diversity while preventing overrepresentation of repetitive motions.}
    \label{supp:tab:data}
    \begin{tabular}{lccc}
        \toprule
        \textbf{Dataset} & \textbf{Sample Rate} & \textbf{Frames} & \textbf{Total Hours} \\
        \midrule
        AMASS & 1/5 & 357,728 & 16.56 \\
        IDEA400 & 1/5 & 424,144 & 19.64 \\
        Other Subsets & 1 & 885,757 & 8.20 \\
        Training Cases & 1/5 & 36,272 & 1.68 \\
        \bottomrule
    \end{tabular}
\end{table}

Our \posetoclip model is engineered specifically for the 15-joint kinematic skeleton used in our animated character, accepting pose inputs represented as $\theta\in\mathbb{R}^{15\times3}$ (Euler angles for each joint). To extend the applicability of our approach to broader research applications, we additionally trained and open-sourced a variant called \posetoclip-M, specifically designed for the standard SMPL skeleton with 24 joints. This model accepts inputs in the form $p=(\theta^{SMPL}, h_{root})$, where $\theta^{SMPL}\in\mathbb{R}^{24\times3}$ encodes the Euler angles of all 24 joints and $h_{root}$ represents the height of the root joint relative to the ground plane.

\subsection{Quantitative Evaluation of \posetoclip}

\posetoclip directly maps humanoid poses to CLIP features, a capability developed by training on over 1.7 million pose-image feature pairs, covering a wide range of actions (\eg, walking, punching, jumping, reclining). To assess the accuracy and completeness of the features learned from this extensive dataset, we construct a similarity matrix $M$.

Each row $i$ of the matrix represents a pose $p_i$, and each column $j$ represents a text description $t_j$. The element $m_{ij} = {\rm sim}(p_i, t_j)$ in the matrix denotes the similarity between $p_i$ and $t_j$. We established the ground truth by recruiting 25 participants to rate the similarity of each pose--description pair on a 0--1 scale. To evaluate whether the method can distinguish different actions under the same text description, we apply a linear transformation to each column so that its minimum and maximum values become 0 and 1, respectively. This normalization process yields the ground truth similarity matrix $M_{GT}$.

We evaluate six approaches using this framework: (i) IsaacGym's native rendering with CLIP, (ii) image prompt with CLIP (which enhances rendered images with a red circle to direct attention), (iii) text prompt with CLIP (which alters subject types in text descriptions), (iv) VLM-RM~\cite{rocamonde2024vlmrm} (which modify CLIP text features to remove agent-specific details), (v) Blender Render with CLIP (utilizing high-fidelity, human-like rendering), and (vi) our proposed model. After undergoing the same normalization process as the ground truth, these methods yield their respective similarity matrices $M_k(k=1,2,...,6)$. We employ matrix similarity $\displaystyle {\rm sim}(M_k, M_{GT}) = 1- \frac{\sqrt{\sum (M_k - M_{GT})^2}}{\sqrt{\sum (M_k^2 + M_{GT}^2)}} $ to score the methods.

The comparison results in \cref{supp:tab:p2c} reveal distinct performance tiers among the evaluated approaches. The four baseline methods—IsaacGym + CLIP~\cite{cui2024anyskill}, Image Prompt, Text Prompt, and VLM-RM~\cite{rocamonde2024vlmrm}—show limited effectiveness with similarity scores tightly grouped between 0.36 and 0.38. This clustering suggests that simple adjustments to rendering or text prompting provide minimal benefit for semantic alignment.

\begin{table}[t!]
    \centering
    \scriptsize
    \caption{\textbf{Comparative Performance Analysis of Pose-to-CLIP Feature Mapping Approaches.} We show matrix similarity scores between human-annotated ground truth and six distinct methods for deriving semantic embeddings from poses. Results demonstrate that while conventional rendering modifications (Image/Text Prompting, VLM-RM) yield minimal improvements over the baseline (IsaacGym + CLIP), photorealistic rendering (Blender + CLIP) achieves substantially higher semantic alignment. Notably, our \posetoclip attains equivalent performance (0.48 vs. 0.49) without requiring any rendering pipeline, validating our direct pose-to-embedding approach as both efficient and semantically accurate. These findings confirm that our model successfully distills the essential pose semantics comparable to high-fidelity visual representation.}
    \label{supp:tab:p2c}
    \begin{tabular}{lcccccc}
        \toprule
        \textbf{Approach} & \textbf{IsaacGym + CLIP} & \textbf{Image Prompt} & \textbf{Text Prompt} & \textbf{VLM-RM} & \textbf{Blender + CLIP} & \textbf{\posetoclip} \\
        \midrule
        \textbf{Matrix Score} & 0.37 & 0.38 & 0.38 & 0.36 & \textbf{0.49} & \textbf{0.48} \\
        \bottomrule
    \end{tabular}
\end{table}

Blender Render with CLIP achieves a score of 0.49 and demonstrates that high-quality visual representation substantially improves pose-text association accuracy. Remarkably, our \posetoclip achieves a nearly identical score of 0.48 while eliminating the rendering pipeline entirely. This performance parity validates our direct mapping approach and confirms that \posetoclip successfully extracts the essential semantic information from physical poses without computationally expensive rendering.

Our final \posetoclip implementation is highly efficient, with just 2.9 million parameters, and achieves an average cosine similarity of 0.86 between predicted and ground truth features on our validation set. This combination of accuracy and computational efficiency makes \posetoclip particularly suitable for real-time applications where rendering would introduce unacceptable latency.

\section{Implementation Details of \model{}}\label{supp:method:RL}

To support reproducibility, we provide comprehensive implementation details for the key components of our framework.

\subsection{\ac{llm} Prompt}

Our systematically engineered \ac{llm} prompt incorporates seven critical components specifically designed to elicit high-quality reward functions:
\begin{enumerate}[leftmargin=*,nolistsep,noitemsep]
    \item \textbf{Role specification:} We position the \ac{llm} as an expert reward engineer specializing in robot learning and motion synthesis, establishing an appropriate technical context for generation.
    \item \textbf{Goal definition:} We explicitly define the objective as designing a reward function for \ac{rl} of a specified task, emphasizing the importance of minimal yet complete formulations.
    \item \textbf{Environment description:} We provide detailed information about the simulation environment, including coordinate system conventions (e.g., up-axis direction), available physical quantities, and relevant state variables accessible to the reward function.
    \item \textbf{Agent specification:} We outline the agent's morphology comprehensively, listing each joint by name and index to enable precise targeting of individual body parts within the reward function.
    \item \textbf{Example template:} We supply a code skeleton demonstrating proper function signature, expected inputs/outputs, and basic structure, serving as a reference pattern for the generated function.
    \item \textbf{Design principles:} We highlight core guidelines for effective reward design, stressing the importance of sparse, well-shaped rewards and noting that auxiliary components already manage stability and locomotion. 
    \item \textbf{Task instruction:} Finally, we include the verbatim natural language instruction $I$, which serves as the primary objective to be optimized by the reward function.
\end{enumerate}

This structured approach significantly improves both the quality and consistency of generated reward functions compared to more generic prompting methods.

\subsection{\posetoclip}

Our \posetoclip implementation combines high-quality visual rendering with an efficient neural architecture design:\begin{enumerate}[leftmargin=*,nolistsep,noitemsep]
    \item \textbf{Rendering pipeline:} We utilize Blender's EEVEE rendering engine, a physically-based real-time renderer, configured to produce 224×224 images that precisely match CLIP's input requirements. The setup incorporates realistic three-point lighting and employs physically accurate materials derived from the SMPL-X model.
    \item \textbf{Neural architecture:} The \posetoclip model features a two-layer MLP with hidden layer dimensions [256, 1024], followed by a direct linear projection to the CLIP feature dimension. Between layers, we incorporate GELU activations to enhance both training stability and generalization capability.
    \item \textbf{Training configuration:} We train the model using Adam optimizer with a learning rate of $10^{-4}$ and a batch size of 512. The training process implements a cosine learning rate schedule with warm-up during the first 10\% of training steps. All experiments run on a single NVIDIA 4090 GPU over eight hours, requiring approximately 18GB of GPU memory.
\end{enumerate}

Despite its compact design of only 2.9 million parameters, the resulting model achieves remarkable fidelity in mapping between pose space and CLIP feature space, demonstrating an average cosine similarity of 0.86 between predicted and ground truth features on our validation set.

\subsection{Low-level Controller}

The hierarchical control system maintains consistent architectural patterns across its key components. The encoder, low-level control policy, and discriminator each employ MLP architectures with identical hidden layer dimensions [1024, 1024, 512]. For efficient representation, we utilize a 64-dimensional latent space $\mathcal{Z}$. 

Critical hyperparameters detailed in \cref{supp:tab:low} include an alignment loss weight of 0.1, uniformity loss weight of 0.05, and a gradient penalty coefficient of 5. We train the low-level controller using PPO~\cite{schulman2017proximal} within the IsaacGym physics simulation environment. The training process runs on a single NVIDIA A100 GPU at a 120 Hz simulation rate over a four-day period, encompassing a diverse dataset of 93 distinct motion patterns to ensure coverage of the humanoid action space.

\begin{table}[ht!]
    \small
    \caption{\textbf{Hyperparameters for the training and operation of our hierarchical control system components.}}
    \label{supp:tab:combined}
    \centering
    \begin{minipage}[t]{0.48\linewidth}
        \centering
        \caption*{\textbf{Low-level Controller} configuration detailing neural architecture parameters, loss weightings, and PPO training settings.}
        \begin{tabular}{ll}
            \toprule
            \textbf{Parameter}         & \textbf{Value}       \\
            \midrule
            dim(Z) Latent Space        & 64       \\  
            Align Loss Weight          & 0.1        \\  
            Uniform Loss Weight        & 0.05      \\  
            $w_{\text{gp}}$ Gradient Penalty   & 5        \\  
            Encoder Regularization     & 0.1      \\  
            Samples Per Update         & 131072   \\  
            Policy/Value Minibatch     & 16384    \\  
            Discriminator Minibatch    & 4096     \\  
            $\gamma$ Discount          & 0.99     \\  
            Learning Rate              & $2 \times 10^{-5}$     \\  
            GAE($\lambda$)             & 0.95     \\  
            TD($\lambda$)              & 0.95     \\  
            PPO Clip Threshold         & 0.2      \\  
            $T$ Episode Length         & 300      \\
            \bottomrule
        \end{tabular}
        \label{supp:tab:low}
    \end{minipage}
    \hfill
    \begin{minipage}[t]{0.48\linewidth}
        \centering
        \caption*{\textbf{Standard \ac{rl} benchmarks.} Benchmark configuration with task-specific architecture dimensions and optimization parameters.}
        \begin{tabular}{ll}
            \toprule
            \textbf{Parameter}         & \textbf{Value}       \\
            \midrule
            ANYmal \& Ant MLP          & [256, 128, 64]\\
            Humanoid MLP               & [400, 200, 100] \\
            ANYmal \& Ant LR           & $3 \times 10^{-4}$\\
            Humanoid LR                & $5 \times 10^{-4}$\\
            Activation                 & elu\\
            $\gamma$ Discount          & 0.99 \\
            KL\_threshold              & 0.008    \\  
            TD($\lambda$)              & 0.95     \\  
            PPO Clip Threshold         & 0.2      \\  
            $T$ Episode Length         & 300      \\
            \bottomrule
        \end{tabular}
        \label{supp:tab:rl}
    \end{minipage}
\end{table}

\subsection{Standard \ac{rl} Benchmarks}

For our standard \ac{rl} benchmark evaluations, we implement control policies using MLPs with a consistent architecture of three hidden layers [400, 200, 100]. These policies are trained using the PPO algorithm~\cite{schulman2017proximal} within the IsaacGym simulation environment. We maintain a learning rate of \(5\times10^{-4}\) throughout the training process and conduct all experiments on a single NVIDIA A100 GPU with a fixed simulation frequency of 60 Hz. Additional hyperparameters governing the training process are fully documented in \cref{supp:tab:rl} for comprehensive reproducibility.

\clearpage

\begin{figure}[ht!]
    \centering
    \begin{overpic}[width=.8\linewidth]{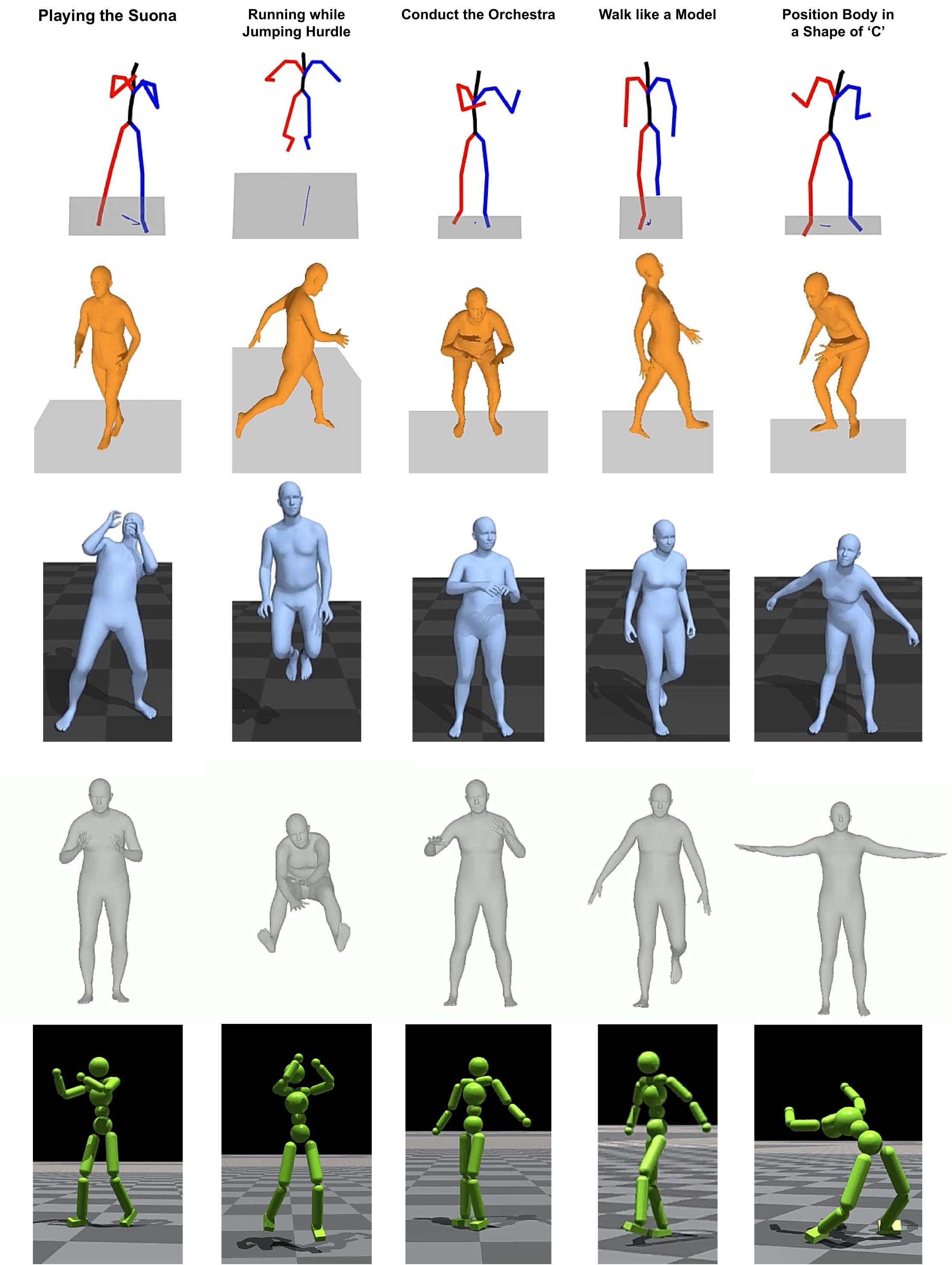}
        \put(34,79){\textbf{(a) TMR}}
        \put(33,61){\textbf{(b) MoMask}}
        \put(32,39.5){\textbf{(c) MotionGPT}}
        \put(31,19.2){\textbf{(d) AvatarCLIP}}
        \put(33,-1.8){\textbf{(e) AnySkill}}
    \end{overpic}
    \vspace{1em}
    \caption{\textbf{Qualitative comparison of motion synthesis approaches.} We compare the performance of five baseline methods across five diverse text prompts. (a) Consistent with our quantitative findings in \cref{tab:comparison}, TMR exhibits minimal capability to interpret the provided instructions. (b) While MoMask performs adequately on common actions like ``jump," it struggles significantly with less familiar prompts. (c) MotionGPT generates motions that partially align with the instructions but predominantly defaults to upright postures with limited movement range, failing to capture the full expressiveness of the intended behaviors. (d) AvatarCLIP demonstrates potential but requires substantial improvement in motion quality. (e) AnySkill (\ac{vlm}) consistently fails to execute the requested actions effectively. For comprehensive evaluation, we provide complete video demonstrations on our project website.}
    \label{supp:fig:baseline1}
\end{figure}

\clearpage

\begin{figure*}[t!]
    \centering
    \begin{overpic}[width=0.75\linewidth]{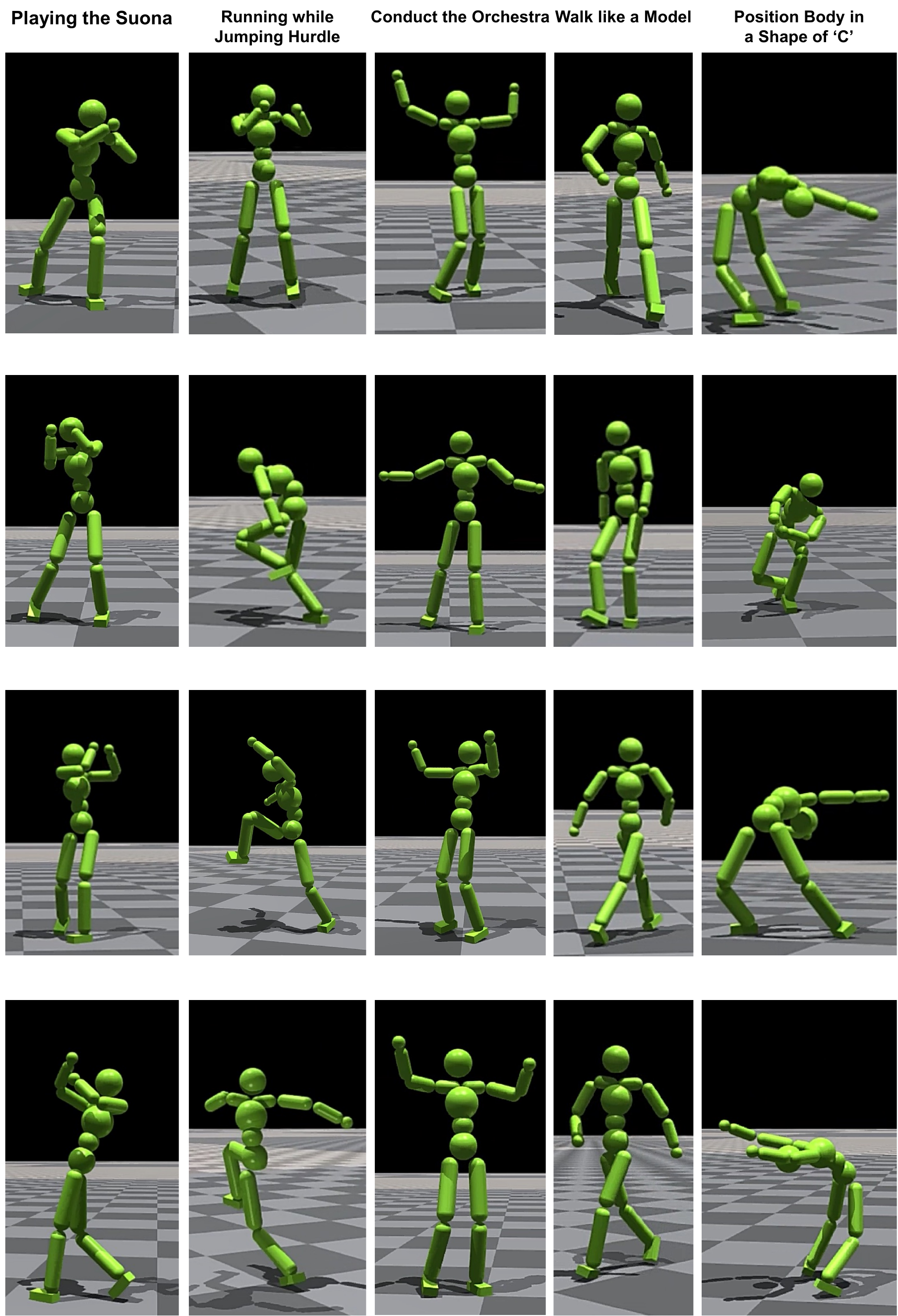}
        \put(29,72.5){\textbf{(a) VLM+LLM}}
        \put(31,48.7){\textbf{(b) LLM}}
        \put(28.5,25.3){\textbf{(c) \posetoclip}}
        \put(25,-2){\textbf{(d) \posetoclip + LLM}}
    \end{overpic}
    \vspace{1em}
    \caption{\textbf{Qualitative results of our motion synthesis approach.} We showcase diverse motion sequences generated by our \posetoclip + \acs{llm} model across five distinct text prompts, demonstrating the system's versatility and expressiveness. Each column showcases a different instruction: ``Playing the Suona'' (leftmost), where the agent adopts appropriate hand positions for the wind instrument with natural body movements; ``Running while Jumping Hurdles,'' exhibiting biomechanically sound preparation, elevation, and landing phases; ``Conduct the Orchestra,'' displaying expressive arm gestures with coordinated body positioning; ``Walk like a Model,'' featuring the characteristic cross-stepping gait rather than simple forward motion; and ``Position Body in a Shape of `C''' (rightmost), showing precise body configuration control. Each row represents a key frame from the continuous motion sequence, highlighting our model's ability to generate temporally coherent, contextually appropriate movements that capture both the semantic meaning and physical nuances of the requested actions. These qualitative results align with our quantitative findings in \cref{tab:ablation}, demonstrating superior motion range, coherence, and task-specific adaptations compared to alternative approaches. Complete motion sequences are available as videos on our project website.}
    \label{supp:fig:baseline2}
\end{figure*}

\endgroup
\end{document}